\documentclass[10pt,twocolumn,letterpaper]{article}

\usepackage[datasets]{wacv} 

\usepackage{CJKutf8}
\newcommand{\zh}[1]{\begin{CJK}{UTF8}{gbsn}#1\end{CJK}}

\definecolor{wacvblue}{rgb}{0.21,0.49,0.74}
\usepackage[pagebackref,breaklinks,colorlinks,allcolors=wacvblue]{hyperref}
\def\wacvPaperID{2220} 
\def\confName{WACV}
\def\confYear{2027}

\title{Beyond BLEU: A Case for Redefining Sign Language Translation Benchmarks}

\author{Oline Ranum, Edward Fish, Simon Hadfield, Richard Bowden\\
Centre for Vision, Speech and Signal Processing (CVSSP), University of Surrey\\
Guildford, GU2 7XH, Surrey, UK\\
{\tt\small \{o.ranum, s.hadfield, r.bowden\}@surrey.ac.uk, ed@tavus.io}
}

\begin{document}
\maketitle
\begin{abstract}
   BLEU-4 is the standard metric for evaluating sign language translation (SLT), but spoken-language metrics may not adequately reflect sign language proficiency. The multimodal, low-resource context of SLT allows models to exploit spurious correlations and spoken-language priors, rather than learning stronger sign representations. In this paper, we evaluate the relationship between spatio-temporal understanding and BLEU-4 across six SLT models on Phoenix-2014T and CSL-Daily, showing that gains in BLEU-4 are not on their own evidence of better sign language understanding. This work introduces an alternative inspired by language-learning assessment, using an open-weight-LLM QA protocol that measures salient content preservation. It aligns more closely with human rankings and is six to seven times more paraphrase-invariant than BLEU-4. Applied to SLT, this protocol targets content transfer, is more robust to train-test overlap, and gives a different picture of the field: the five gloss-free systems are largely within noise of one another on Phoenix-2014T, while the gloss-supervised system stands $9.3$ points higher, a gap invisible to BLEU-4.
\end{abstract}
    
 \section{Introduction}
        \label{sec:intro}
        In language learning, evaluation rarely hinges on the exact reproduction of phrasing. Instead, it asks what was conveyed: who acted, what happened, and under what conditions. Conversely, SLT performance is still measured almost entirely by BLEU \cite{papineni2002bleu}, an n-gram precision metric that quantifies how closely a system reproduces a reference. This definition of quality may be adequate at scale, but SLT remains a low-resource domain in which datasets cover only narrow linguistic scopes. Consequently, strong spoken language models can achieve high BLEU scores by exploiting regularities in the target-language distribution alone, while paying little attention to the visual language signal.
    
    The modalities also structure information differently: signing encode simultaneously and densely, using space for placement, depiction, and constructed action, while many spoken-language grammatical elements are absent (e.g., articles: \emph{the}, \emph{a}, \emph{an}; adpositions: \emph{for}, \emph{to}). The resulting source-target mapping is highly non-isomorphic, producing a large space of valid translations and reducing the reliability of BLEU. Metrics that evaluate content preservation may better reflect this reality, tolerating varied linguistic realization while rewarding preservation of salient information.
    
    This paper shows that BLEU fails to reliably reflect improvements in sign understanding: BLEU-4 retains substantial scores despite strong input corruptions, and disproportionately rewards components with low correlation to the visual language signal. To address these issues, this paper proposes a question-answering (QA)-based evaluation framework which is akin to how human comprehension is tested.
    The metric is first validated on spoken-language benchmarks, assessing its sensitivity to semantic shifts and its alignment with human judgment. Then it is applied to the six state-of-the-art SLT models and two sign language benchmarks, demonstrating its scalability and efficiency, and is less influenced by train-test similarities. These findings lead to a revision of current performance rankings.

\section{Background \& Related Work}
\subsection{Sign Language Translation}

Sign language modeling spans recognition (SLR) \cite{li2020, RASTGOO2021, Huang2018, ranum2026, Zuo2022, raude2024, ranum2024}, production (SLP) \cite{Zhengdi2024, Xie2023, yin2024, wang2025, xia2026} and translation (SLT) over video or pose inputs paired with gloss or text supervision \citep{bragg2019sign, desai2024systemic, fox2023best, brown2026}. Glosses are written labels for individual signs in citation form, which capture a limited portion of the signed content.
Early SLT systems predicted gloss sequences before generating text \citep{chen2023twostream, chen2023simple, yin2021simulslt, zhou2021improving, zhou2022spatial, zhang2023sltunet}; whereas more recent works have explore gloss-free approaches which map visual inputs directly to text \cite{Duarte2021, Albanie2021, tanzer2025, shi2022, uthus2023} with end-to-end transformers and large pretrained language decoders \citep{wong2024sign2gpt, kim2025leveraging, chen2025c2rl, zhou2023glossfree, asasi2025beyond, Sincan2025}.

\paragraph{Evaluation, grounding and interpretability} Datasets used for SLT are currently very limited. The most commonly used dataset Phoenix14T \cite{Camgoz_2018_CVPR} contains just 8,257 German Sign Language videos from 9 interpreters. Another frequent dataset is CSL-Daily \cite{Zhou2021}, which contains 20,654 videos over 2,000 daily-life phrases. Both datasets are recorded in controlled settings, limiting domain variability, signer diversity, and naturalness. We use them here because they are the only SLT benchmarks with gloss annotations, offering a proxy for the signed content in the attribution analysis of Section~\ref{sec:bleulimits}. We evaluate six models reporting state-of-the-art results: GFSLT-VLP \cite{zhou2023}, FLA-LLM \cite{chen2024}, CiCo \cite{Yiting2023}, SignCL \cite{Jinhui2024} and C2RL \cite{Chen2024a}, all gloss-free, together with the single-stream (RGB-only) variant of the gloss-supervised TwoStream \cite{chen2022}; architectures and checkpoints are presented in Appendix~\ref{app:datasets}.

While BLEU-4 remains the de facto SLT metric, several recent works have challenged its adequacy \cite{cory2026, brown2026, jiang2025, Saunders2020}. SignBLEU \citep{kim2024} extends BLEU to a multi-channel formulation capturing manual and non-manual components, but is restricted to linguistic corpora with rich annotations. For gloss-free evaluation, \citet{hamidullah2026} find that reduced visual dependence correlates with higher hallucination rates, while \citet{alakin2026} show that BLEU is disproportionately influenced by test examples resembling the training distribution.

    
\subsection{QA-based and LLM-based Evaluation}
LLM-based QA evaluation has proven a viable alternative to established spoken language translation metrics \citep{ki2025, han2022, krubinski2021}; \citet{Kamalloo2024} show that open instruction-tuned LLMs correlate competitively with human judgment. \citet{fernandes2025} extend this to MT with TREQA, an LLM-based QA framework for paragraph-level translation that outperforms or is competitive with chrF, COMET, COMET-QE, GEMBA-MQM, MTEQA. \citet{zhang2025} propose LiTransProQA, a QA framework for literary translation that incorporates professional translators' assessment criteria. In this work, we apply the LLM-based QA framework to evaluate SLT. We show that it aligns more closely with human ranking than BLEU-4, and that its scores are far less inflated by a test reference's resemblance to the training data.

        \section{Preliminaries: Correlations and Attributions of BLEU-4 in SLT}
        \label{sec:bleulimits}
        In multimodal translation, target-side metrics such as BLEU do not provide insight into whether improvements reflect stronger visual understanding, spoken language modeling or overfitting. This section identifies reward structures and dependencies that limit BLEU's ability to reflect sign proficiency.

      \subsection{Robustness to Input Corruption}
    
\begin{figure}[!t]
            \centering
            \includegraphics[width=\linewidth]{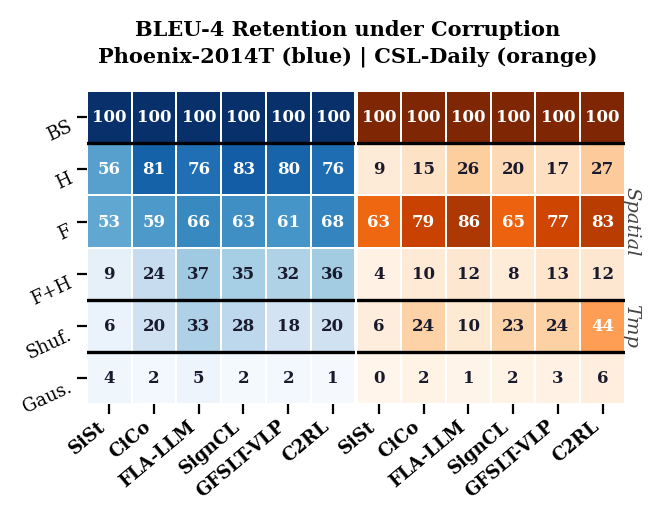}
            \caption{BLEU-4 retention (\% of baseline) under input corruptions: spatial masking of hands (H) and face (F), temporal shuffling of frames (Shuf) and Gaussian noise.}
            \label{fig:input_modification}
        \end{figure}

        Figure \ref{fig:input_modification} shows BLEU-4 retention scores under spatial and
        temporal corruption. Despite these interventions, the models often maintain strong performance, suggesting a reliance on target-side priors, memorization, or other spurious correlations (per-condition details in Appendix~\ref{app:inputcorruption}). Replacing the visual representation with noise leaves every model at $1.2$ BLEU-4 points or below, fixing the floor a system reaches while reading nothing. If the score were a true reflection of
        sign comprehension, masking the signer's hands and face or shuffling the frames should approach that floor. However, retention stays well above the noise floor throughout: masking a single articulator leaves most models above half their baseline, and even with both articulators masked or the frames shuffled, scores remain several times, up to thirty times, the noise baseline. Masking both hands leaves CiCo, SignCL and GFSLT-VLP at $81\%$, $83\%$ and $80\%$ of baseline on Phoenix-2014T, where masking the face leaves every model at $53$--$68\%$, and between $63$--$86\%$ on CSL-Daily.
        
        These gaps matter because of their size relative to the margins in use. On Phoenix-2014T the five gloss-free systems retain $7.3_{\pm 0.9}$ BLEU-4 points
        with both articulators masked, close to three times the $2.5$-point spread that
        separates the six systems at baseline (Table~\ref{tab:combined_results}). What a
        model scores while reading none of the signing therefore exceeds the differences
        the field uses to rank systems, so a BLEU-4 gain at the current state of the art
        is not on its own evidence of better sign proficiency.


        \subsection{Modalities \& Award Structures}
        As introduced in Section~\ref{sec:intro}, the non-isomorphic nature of SLT shapes how evaluation should be contextualized. No current annotation scheme offers a precise account of what is articulated in the sign space, but gloss may offer a first approximation. Figure~\ref{fig:bleuretention} pairs gloss and text-token POS distributions with the corresponding BLEU-4 attribution: for both Phoenix-2014T and CSL-Daily, averaged over six systems, $37\%$ of BLEU-4 is attributable to function words occurring far less frequently in gloss than in text (German adpositions, 57 gloss vs.\ 1155 spoken tokens, 3.65 points; Chinese particles, 33 vs.\ 1279, 1.81 points).
        
    \section{Implications for SLT Evaluation}
\begin{figure}[!b]
        \centering
            \includegraphics[width=0.88\linewidth]{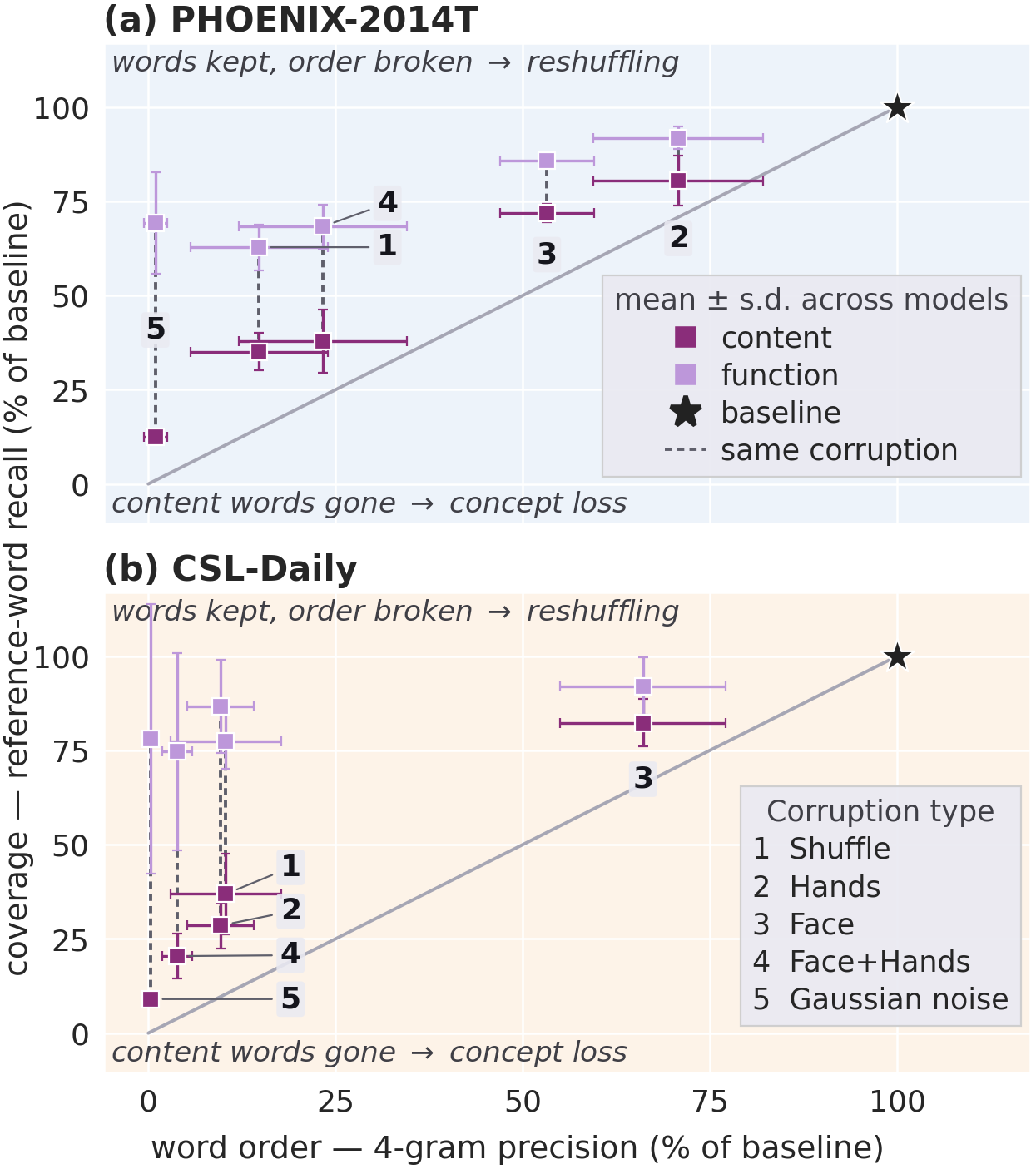}
        \caption{What survives corruption. Each condition contributes one content and
        one function marker (mean $\pm$ SD over models), placed by retained
        4-gram precision (word order) vs.\ retained reference-word recall
        (coverage), both as \% of baseline.}
            \label{fig:structdiff}
        \end{figure}

    Section~\ref{sec:bleulimits} shows that BLEU-4 often rewards structures with limited or absent correspondence to the visual language signal, by a margin larger than separates current systems. Plausibly, this reflects overfitting driven by narrow linguistic scope.  Figure~\ref{fig:structdiff} separates output surviving corruption along two axes: whether it preserves the reference's words or only their order, and whether those words are content or function. This distinction matters because the two categories are not equally recoverable from target-side priors alone: function words occupy a narrow, concentrated vocabulary, while content words are broad and long-tailed (Appendix~\ref{app:langdist}).

    Two patterns hold in both datasets under every condition: coverage is retained better than 4-gram precision, and function-word coverage is retained far better than content-word coverage. What survives corruption is thus predominantly what target-side priors can supply, exactly what a metric rewarding explicit function-word forms is most exposed to. We propose instead that reward functions target the preservation of salient content. This framework has three immediate benefits: First, such a metric tolerates paraphrasing. Second, it discounts surface-level structure that survives losing the visual signal. Third, because credit is awarded almost entirely for content, a drop under occlusion reads as content lost rather than correct word order.

    \subsection{Problem Formulation:\\Salient Content Preservation Metrics}
    \label{sec:problem}

    Receptive assessment in early-stage sign language learning \cite{Hauser2016, Rosenburg2020} suggests a natural way to apply these perspectives: probe comprehension by querying the conveyed content, and assess whether the responses recover the salient information of the input. The above leads to three central questions:
    
    \begin{enumerate}[noitemsep]
        \item How can the preservation of salient content be measured reliably and at scale?
        \item Does content-based evaluation reduce the risk of overestimating systems that exploit target-side regularities?
        \item How effectively do current models preserve the salient content of the source discourse?
    \end{enumerate}
    
    \noindent The remainder of the paper addresses these in turn. Section~\ref{sec:qatext} implements the first question as an LLM-based QA protocol, evaluated on single-modality benchmarks. Section~\ref{sec:qaslt} then transfers the protocol to SLT, addressing the second and third questions.
    
            \begin{figure}[!b]
            \centering
            \includegraphics[width=0.88\linewidth]{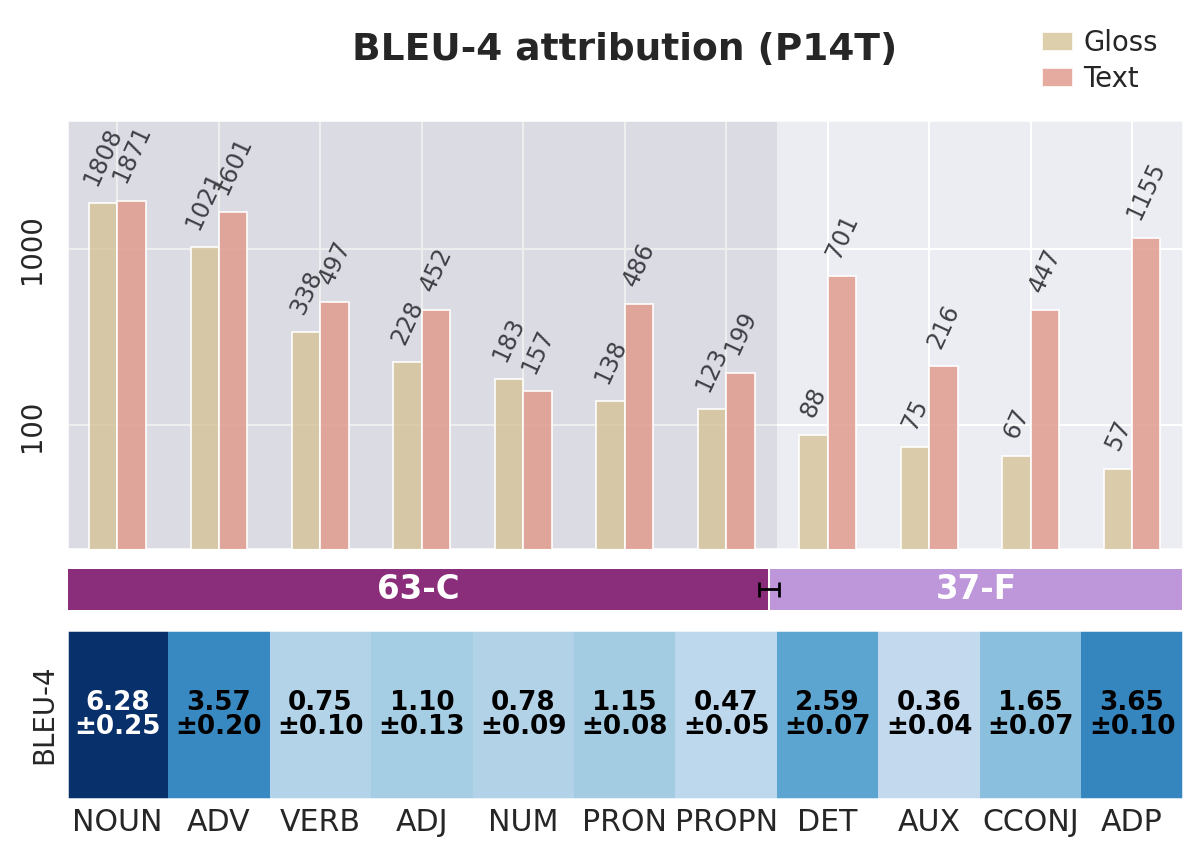}\\[0.6em]
            \includegraphics[width=0.88\linewidth]{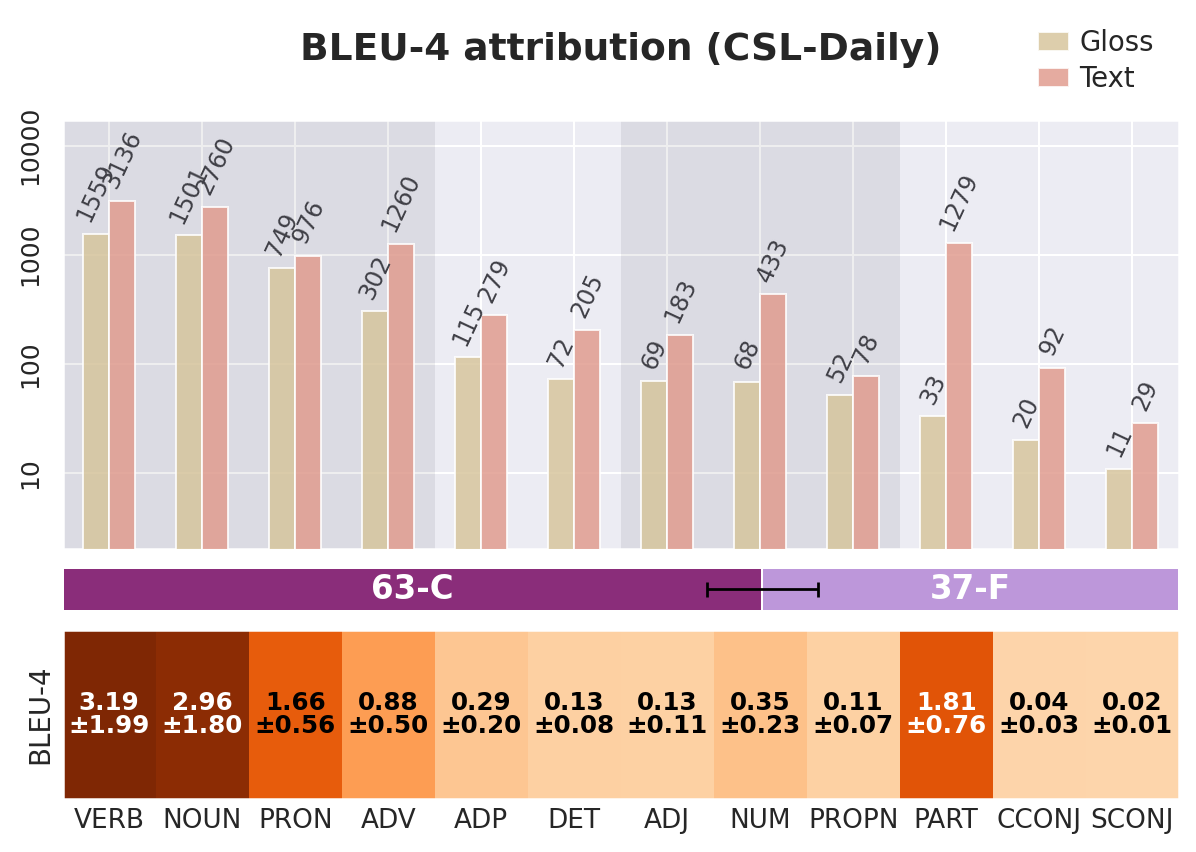}
            \caption{Gloss and text POS distributions with BLEU-4 attribution, for Phoenix-2014T (top) and CSL-Daily (bottom).}
            \label{fig:bleuretention}
        \end{figure}

        \clearpage
        \section{Measuring Content Preservation}
        \label{sec:qatext}
        A metric for salient content preservation should ask what a translation conveys rather than how it is worded, which sets two requirements: invariance to wording at fixed content, and sensitivity to changes in the content itself. This section implements such a protocol as an LLM-driven question generation and answering pipeline, then establishes its behavior on spoken-language benchmarks, 
        where we have human annotations available.

        \subsection{Evaluation Protocol}
        \paragraph{Task definition} Given a human reference translation $r$, the protocol builds a bank $Q_s$ of quality-controlled multiple-choice questions and scores a model prediction $s$ by the fraction of $Q_s$ it can answer correctly from $s$ alone (Appendix~\ref{app:metricdefs}), using an open-weight LLM served via vLLM \cite{kwon2023}. Implementation details and all prompts are provided in Appendix~\ref {app:prompts}.
    
        \paragraph{QA-Bank Generation}
        The QA-bank generation is performed in three main phases. To increase coverage, the LLM is prompted to extract all content units from $r$ related to nine categories (entity, action, attribute, quantity, time, location, relation, negation, polarity). In phase 2, the LLM is prompted to generate questions about that unit. $|Q_s|$ thereby scales with the reference's semantic density; per-language yields
        and lemma coverage are in Appendix~\ref{app:qayield}. In phase 3, distractors related to each unit and question are generated independently. Four per question, each plausible but incorrect, mutually distinct, non-paraphrastic, and matched in language and register. The five options are deterministically permuted and a language-specific \emph{not stated} sentinel fixed in a sixth position, which lets the answerer decline rather than guess and floors the metric on non-equivalent content rather than at chance ($1/6$).
    
    
        \paragraph{Quality control}
        Three LLM-executed gates filter the assembled items. A \emph{round-trip} gate requires the answerer, shown the reference itself, to select the correct option, since an item unanswerable from the reference cannot fairly be asked of a candidate. A \emph{world-knowledge probe} presents the question with an empty passage and rejects items still answered correctly, removing those answerable from language priors alone (cf.\ Section~\ref{sec:bleulimits}). An \emph{ambiguity} gate re-presents the reference, the question, and the five
        content options in a separate pass, asks which of them the reference states or clearly implies, and rejects the item if more than one does. Candidates are then scored from the passage, question, and options alone. Human verification of every stage is reported in Appendix~\ref{app:humaneval}.

        \paragraph{Yield and content coverage}
        Table~\ref{tab:questiongeneration} reports bank sizes and content reach. Question counts scale with sentence complexity as intended: short colloquial subtitles (OpusParcus) yield 3.2 questions per reference against 11.1 for the longer Wikipedia-derived sentences of PAWS-X. Since the metric can only penalise a candidate for content its questions probe, reach is quantified using the POS machinery of Section~\ref{sec:bleulimits}: \emph{content coverage} is the fraction of a reference's content lemmas that reappear in the QA items generated from it, and the bank reaches about $88.6\%$ on PHOENIX14T and $94.6\%$ on CSL-Daily.

            \begin{table}[!h]
            \centering\small
            \setlength{\tabcolsep}{2.5pt}
            \begin{tabular}{llrcc}
    \toprule
        \textbf{Benchmark} & \textbf{L} & $N_{\text{ref}}$ & $\bar N_Q$ & Cov.\ (\%)\\
    \midrule
        \multicolumn{5}{l}{\emph{Signed}} \\
        PHOENIX14T & 1 & 641 & $9.15_{\pm 0.05}$ & $88.6_{\pm 0.19}$  \\
        CSL-Daily  & 1 & 1{,}175 & $6.78_{\pm 0.03}$ & $94.6_{\pm 0.11}$ \\
    \midrule
        \multicolumn{5}{l}{\emph{Spoken}} \\
        WMT 2011--2024 & 5 & 15{,}938 & $12.1$ & $92.8$ \\
        OpusParcus & 6 & 19{,}492 & $3.2$ & $84.9$ \\
        PAWS-X & 7 & 13{,}862 & $11.1$ & $95.0$  \\
    \bottomrule
    \end{tabular}
        \caption{QA yield and coverage pooled over $L$ languages. $\bar N_Q$ = average number of questions per reference. \emph{Cov.} (\%) = share of the reference's content lemmas covered by the QA bank. Subscripts are standard deviations over ten independent bank generations.}
        \label{tab:questiongeneration}
        \end{table}

        \subsection{Validation}
        \label{sec:qavalidation}
            To assess robustness to meaning-shift and wording, the QA pipeline is evaluated
            on four benchmarks: paraphrase invariance (WMT19-P \cite{freitag2020};
            WMT21-multiref \cite{akhbardeh2021,freitag2021}); sensitivity to shifts in
            meaning from partial paraphrases (OpusParcus \cite{creutz2018}); sensitivity
            to flipped meaning under fixed wording (PAWS-X \cite{yang2019}); and
            alignment with human MT-system rankings over the WMT campaigns from 2011 to
            2024 \cite{callison2011, callison2012, bojar2013, bojar2014, bojar2015, buckkoehn2016, bojar2017, freitag2021}. Dataset details are provided in
            Appendix~\ref{app:datasets}, and per-language results are provided in
            Appendix~\ref{app:validation}.

\begin{figure}[!b]
            \centering
            \includegraphics[width=\linewidth]{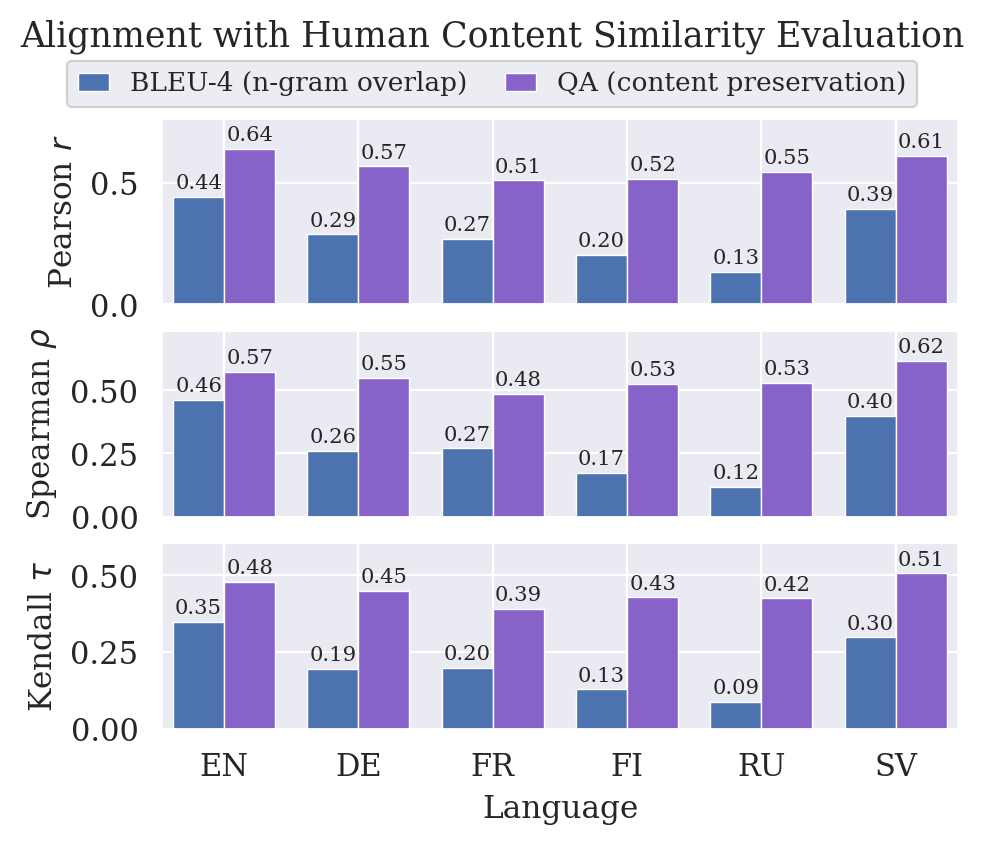}
            \caption{Correlation of BLEU-4 and QA with human paraphrase-quality ratings on
            OpusParcus per language (\texttt{qwen2.5-32b}). Pearson $r$, Spearman
            $\rho$ and Kendall $\tau$ over all graded pairs.}
            \label{fig:oprank}
        \end{figure}

\begin{figure}[!b]
            \centering
            \includegraphics[width=\linewidth]{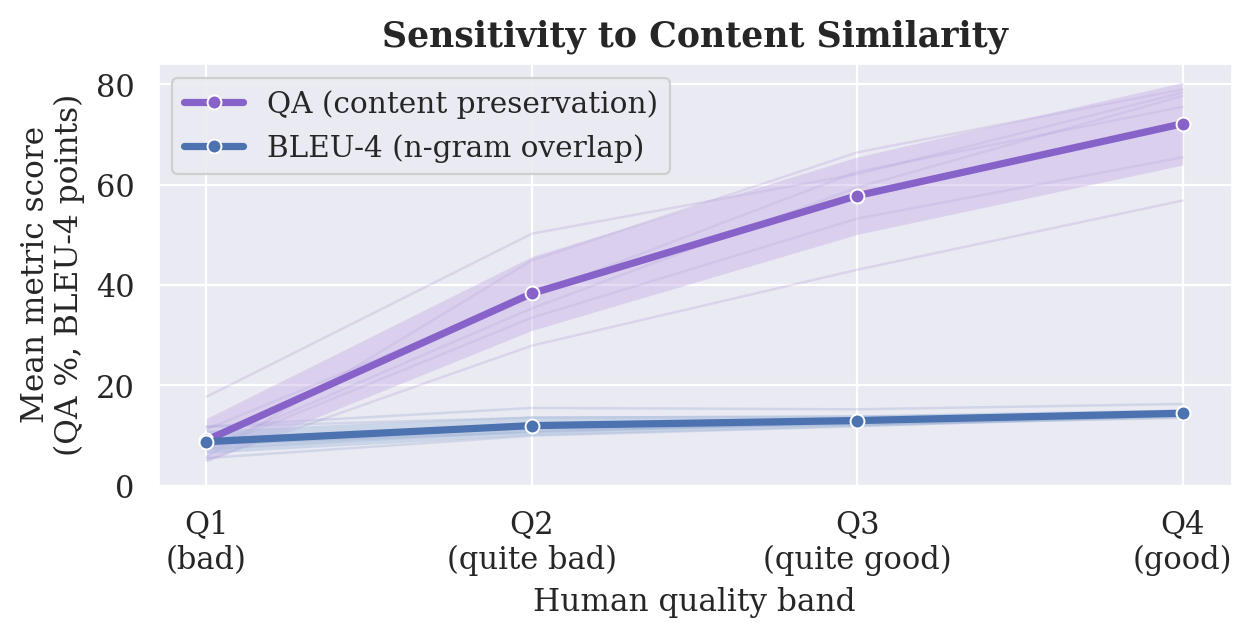}
            \caption{Mean metric score by human similarity band on OpusParcus (Q1 = low similarity, Q4 = clean paraphrase), mean $\pm$ SD over the six languages. 
            }
            \label{fig:opsim}
        \end{figure}

        \paragraph{Invariance and Sensitivity}        
        Each WMT group collects renderings of one source sentence that differ in wording but not content: one supplies the reference from which questions are generated, the rest are scored against it, and renderings of other sentences supply a floor (Appendix~\ref{app:metricdefs}). Invariance is the gap to that floor divided by the fluctuation within a group: on WMT19 BLEU-4 fluctuates by $7.7$ points against a $16.5$-point gap, an SNR of $2.1$, where QA reaches $12.5$; on WMT21 the two are $3.1$ and $23.0$ (Table~\ref{tab:validation}). The two collections differ in provenance, deliberate paraphrasing against independent translation, so their agreement is not a protocol artefact. Sensitivity is tested on OpusParcus, whose ratings are graded: QA tracks the human scale on average twice as closely as BLEU-4 in every language (Figure~\ref{fig:oprank}), rising from $9$ to $72$ across the four quality bands where BLEU-4 moves from $9$ to $14$ (Figure~\ref{fig:opsim}).

        \paragraph{Meaning inversion}
        PAWS-X pairs keep surface overlap high whether or not meaning is inverted, so lexical overlap is uninformative by design. Scoring one sentence of each pair against the other yields two populations of scores, one for true paraphrases and one for meaning-flipped pairs, and we summarise how far apart they lie by ROC-AUC: the probability that a randomly drawn true paraphrase outscores a randomly drawn flipped pair. A metric blind to the inversion sits at $0.5$ and one that separates the two populations perfectly at $1.0$. Being threshold-free, it compares BLEU-4 and QA on a common footing despite their different scales. QA performs better than BLEU-4 in all seven languages, $0.83$ against $0.61$. That QA still awards flipped English pairs $55.7\%$ against $95.4\%$ for true paraphrases is expected: PAWS-X alters one or two arguments, so most questions stay answerable.

        \begin{table}[!b]
        \centering\small
        \setlength{\tabcolsep}{5pt}
            \begin{tabular}{llrr}
    \toprule
    \textbf{Benchmark} & \textbf{Statistic} & \textbf{BLEU-4} & \textbf{QA} \\
    \midrule
    WMT19 \small(invariance)    & SNR             & 2.1   & \textbf{12.5} \\
    WMT21 \small(invariance)    & SNR             & 3.1   & \textbf{23.0} \\
    OpusParcus \small(graded)   & $\bar\rho$      & 0.28  & \textbf{0.55} \\
    PAWS-X \small(inversion)    & AUC             & 0.61 & \textbf{0.83} \\
    \bottomrule
    \end{tabular}
        \caption{Robustness: SNR $=\Delta/\bar\sigma$ over
        meaning-equivalent groups (higher = more invariant); $\bar\rho$ is Spearman
        correlation with human paraphrase-quality ratings; AUC is $P(\text{score$_{paraphrase}$} > \text{score$_{flipped}$})$,
        averaged over seven PAWS-X languages. QA performs better than BLEU-4 across all metrics.}
        \label{tab:validation}
        \end{table}

        \begin{figure}[!b]
            \centering
            \includegraphics[width=\linewidth]{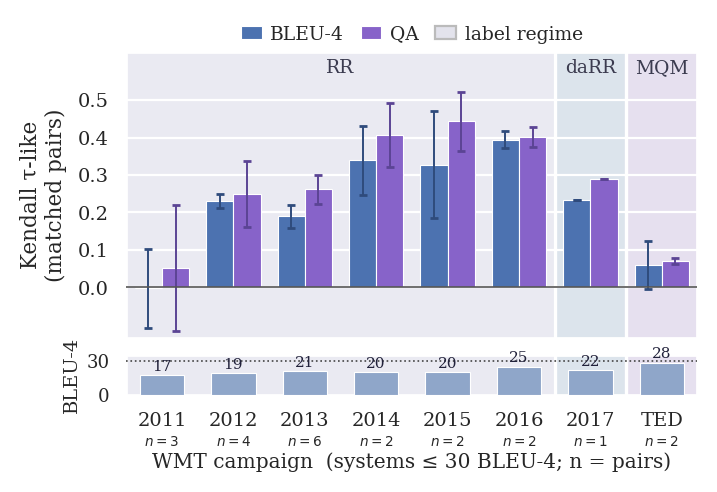}
            \caption{Kendall $\tau$-like concordance with human judgments, over WMT campaigns restricted to systems at or below 30 BLEU-4. QA is more aligned with human ranking than BLEU-4 in all but one campaign.}
            \label{fig:wmt}
        \end{figure}

        \paragraph{Human rankings of MT systems}
        Current text-only MT is near human performance, so its error profiles differ from those of SLT systems; restricting the WMT campaigns to systems averaging below 30 BLEU-4 gives a first approximation. Figure~\ref{fig:wmt} reports Kendall concordance with human ranking. QA matches or exceeds BLEU-4 in every campaign, though the intervals overlap in most; separation is clearest in 2013--2015, where systems are weak enough for content transfer to differ between them but not so weak that both metrics floor.

        \section{QA-based Evaluation for SLT}
        \label{sec:qaslt}
            The spoken-language benchmarks establish that the protocol behaves as a
        content metric should. This section applies it to the six SLT models, addressing
        the three evaluation questions of Section~\ref{sec:problem} in order.

    \subsection{Efficiency and Stability}
    \label{sec:cost}
    The first question asks how salient content preservation can be measured reliably and at scale.
    

    \paragraph{Cost} Costs depend on deployment, so each stage is reported separately. Each experiment is conducted on a single RTX~5090 with \texttt{qwen2.5-32b} served under vLLM (Table~\ref{tab:cost}). Bank construction, encompassing content extraction, two QA-generation passes and three quality-control gates, is the expensive stage at $22$--$24$ minutes per benchmark. Answering, one short-prompt call per admitted question, is up to an order of magnitude cheaper, putting the recurring cost of evaluating an additional system at roughly $2$--$3$ GPU minutes.
        
    \begin{table}[!b]
    \centering\small
    \setlength{\tabcolsep}{3pt}

    \begin{tabular}{lrrrr}
    \toprule
    \textbf{Generation} & $N_{\text{ref}}$ & $|Q|$ & \textbf{wall (s)} & \textbf{GPU-h} \\
    \midrule
    Phoenix2014T & 642 & $5{,}863_{\pm32}$ & $1{,}420_{\pm11}$ & $0.394_{\pm0.003}$ \\
    CSL-Daily     & 1{,}176 & $7{,}967_{\pm41}$ & $1{,}319_{\pm5}$ & $0.367_{\pm0.002}$ \\
    \midrule
    \textbf{Answering} & \textbf{calls} & \textbf{ans/s} & \textbf{wall (s)} & \textbf{GPU-h} \\
    \midrule
    Phoenix2014T & 5{,}863 & $48.8_{\pm0.1}$ & $120_{\pm1}$ & $0.033_{\pm0.001}$ \\
    CSL-Daily     & 7{,}967 & $46.0_{\pm0.2}$ & $173_{\pm1}$ & $0.048_{\pm0.001}$ \\
    \bottomrule
    \end{tabular}

    \caption{Computational cost. Generation costs are averaged over 10 bank-construction runs, and answering costs over 60 answering passes, per benchmark. $|Q|$ is the number of questions admitted by the quality-control gates, out of $10{,}030_{\pm61}$ candidates for Phoenix-2014T and $12{,}373_{\pm48}$ for CSL-Daily.}
    \label{tab:cost}
    \end{table}
    
        \paragraph{Stability} A metric is only useful if repeating an experiment yields
        the same result. Rerunning the generation pipeline
        does not necessarily reproduce a bank. Roughly three-quarters of the admitted
        items recur between two banks, partly by design, since the second generation pass
        samples at a non-zero temperature (Appendix~\ref{app:stability}). However, that
        variability does not carry through to the scores. Table~\ref{tab:stability}
        reports ten end-to-end repeats, each regenerating content units, questions and
        answers against fixed system outputs. Across those repeats a system's QA accuracy
        moves by $\sigma = 0.17$--$0.34$ percentage points on Phoenix-2014T and
        $0.11$--$0.26$ on CSL-Daily. Holding the bank fixed and bootstrap-resampling the
        test instances instead moves it by $2.4$--$2.7$ and $1.4$--$2.0$ points
        (half-width of the $95\%$ interval; Appendix~\ref{app:ci}), between seven and
        sixteen times more for every system. Which questions get asked therefore matters
        far less than which sentences the test set happens to contain. The banks are
        correspondingly stable: their size varies by about half a percent, and every run
        covers at least $99.8\%$ of references, varying by at most one reference. A QA gap
        below roughly half a point is therefore within run-to-run noise and should not be
        read as a difference between systems.

        \begin{table}[!b]
            \centering\small
            \setlength{\tabcolsep}{4pt}
            \begin{tabular}{lrrr}
            \toprule
            \textbf{Quantity} & \textbf{mean}$_{\pm\sigma}$ & \textbf{range} & CI/2 \\
                \midrule
            \multicolumn{4}{l}{\emph{PHOENIX-2014T, QA \% per system}} \\
            C2RL         & $53.05_{\pm 0.32}$ & 52.56--53.47 & 2.65 \\
            CiCo         & $56.47_{\pm 0.34}$ & 56.04--57.03 & 2.57 \\
            Fla-LLM      & $55.23_{\pm 0.17}$ & 54.99--55.48 & 2.53 \\
            GFSLT        & $53.93_{\pm 0.24}$ & 53.51--54.29 & 2.50 \\
            SignCL       & $54.28_{\pm 0.17}$ & 53.93--54.54 & 2.60 \\
            SingleStream & $65.85_{\pm 0.27}$ & 65.28--66.22 & 2.39 \\
            \addlinespace
            bank size $|Q|$    & $5{,}863_{\pm 32}$ & 5{,}803--5{,}906 & \\
            refs.\ covered     & $641.0_{\pm 0.0}$ & 641--641 & \\
            \midrule
            \multicolumn{4}{l}{\emph{CSL-Daily, QA \% per system}} \\
            C2RL         & $11.55_{\pm 0.12}$ & 11.35--11.72 & 1.38 \\
            CiCo         & $26.89_{\pm 0.14}$ & 26.65--27.13 & 1.73 \\
            Fla-LLM      & $40.95_{\pm 0.21}$ & 40.57--41.26 & 2.03 \\
            GFSLT        & $22.19_{\pm 0.11}$ & 22.00--22.32 & 1.78 \\
            SignCL       & $21.09_{\pm 0.17}$ & 20.85--21.45 & 1.70 \\
            SingleStream & $61.42_{\pm 0.26}$ & 60.93--61.79 & 2.01 \\
            \addlinespace
            bank size $|Q|$    & $7{,}967_{\pm 41}$ & 7{,}893--8{,}037 & \\
            refs.\ covered     & $1{,}175.2_{\pm 0.4}$ & 1{,}175--1{,}176 & \\
            \bottomrule
            
            \end{tabular}
            
                \caption{Accuracies over ten runs, each with content units, questions, and system answers regenerated. $\sigma$ is variation across question banks; CI/2 is the mean half-width of the within-bank 95\% bootstrap interval(Appendix~\ref{app:ci}).}
            \label{tab:stability}
        \end{table}

        \subsection{Exposure to Target-Side Regularities}
        \label{sec:attribution}
        \label{sec:sensitivity}
    
        The second question asks whether content-based evaluation reduces the risk of overestimating systems that exploit target-side regularities. This is influenced by what the metric awards, and how much it rewards test examples that resemble the training data.

        \paragraph{Award structures} Under the content/function POS division of Figure~\ref{fig:bleuretention}  (tagging and attribution in Appendix~\ref{app:pos}), BLEU-4 derives 37\% of its score from function words, whereas QA spends $95\%$ on content for Phoenix-2014T and over $99\%$ for CSL-Daily. Function words are also what best survives corruption. Replacing the visual signal with noise leaves function-word recall at $69\%$ of baseline on Phoenix-2014T and $78\%$ on CSL-Daily, while content-word recall falls to $7$--$16\%$ (Figure~\ref{fig:structdiff}). A third of BLEU-4 is therefore committed to a word class the model can still produce once the signal the metric is meant to measure is gone. This opens a channel for improvement without comprehension, where a system can raise BLEU-4 by modeling the target language better in the very category the signed source largely does not encode, indistinguishably from a gain earned by reading the signing. Nor is the channel narrow, corresponding to approximately $8$ BLEU-4 points on Phoenix-2014T against the $2.5$ points that separate the six systems. QA leaves almost none open: with $95\%$ or more of its credit on content.

       \paragraph{Exposure to train--test overlap}
    
        \citet{alakin2026} show that BLEU scores a test sentence higher when it resembles
        the training data, crediting recall of the training targets as though it were
        translation quality, and doing so most where the training distribution is narrowest,
        which is the situation in SLT. Whether the QA score depends on that resemblance is therefore worth measuring.
    
        Following \citet{alakin2026}, we characterize each test instance by a \emph{training-likeness}
        $\ell_i = \max_{t \in T_d} \operatorname{sim}(r_i, t)$, the character-level similarity
        between its reference and the closest training target, so $\ell = 100$ means the
        reference occurs verbatim in training. Within a sliding window over $\ell$ we regress the per-instance score on $\ell$ at fixed reference length, and express the slope as a percentage of that model's own mean score: a value of $10$ says one further likeness point buys ten percent of the score the system reports, and $0$ says the metric is indifferent to how training-like a sentence is. Normalizing this way is what makes the two metrics comparable. On Phoenix-2014T BLEU-4 averages about $22$ and QA about $56$, so the same raw slope would mean very different things for each; dividing by each metric's own mean removes that difference in scale and lets both be read on a single axis. Window widths, the treatment of exact copies and a cumulative view are in Appendix~\ref{app:overlap}.
    
\begin{figure}[!t]
            \centering
            \includegraphics[width=\linewidth]{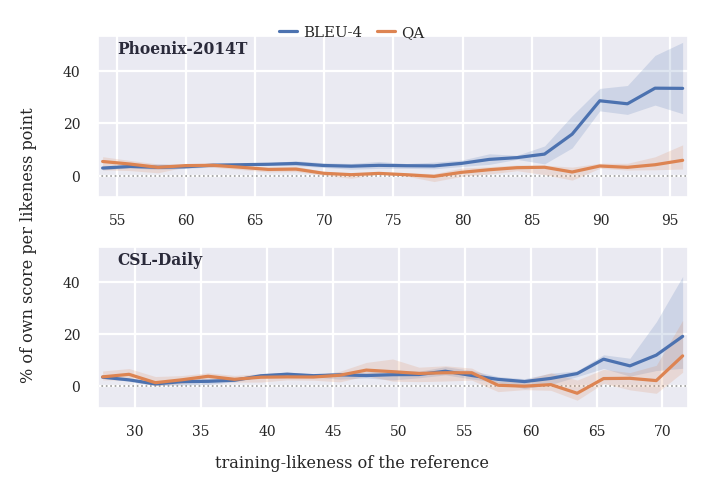}
            \caption{Local sensitivity of BLEU-4 and QA to training-likeness $\ell$, at
            fixed reference length. Curves are the mean over $M=6$ models; bands span
            their min--max. Exact duplicates ($\ell=100$) are excluded as a point mass,
            and, on Phoenix-2014T, because their references are shorter than
            neighbouring points. }
            \label{fig:sensitivity}
        \end{figure}

        On Phoenix-2014T (Figure~\ref{fig:sensitivity}) BLEU-4 sits at $3.9\%$ per
        likeness point over most of the range, below $\ell = 80$, then rises eightfold to
        $32\%$ for the most training-like sentences, where QA moves only from $2.4\%$ to
        $6.5\%$. The exact copies, excluded from the curves, show the same asymmetry in
        levels rather than slopes: on the $31$ references that occur verbatim in training,
        sentence BLEU-4 averages $52.5$ against $21.4$ elsewhere, a $2.4$-fold jump, where
        QA averages $79.7$ against $57.1$, a $1.4$-fold one. CSL-Daily reproduces the
        shape more weakly: over the top quarter of its likeness range BLEU-4 averages
        $8.4\%$ per likeness point against $2.5\%$ for QA, a $3.4$-fold gap where the
        Phoenix-2014T figures of $23.5\%$ and $3.6\%$ give $6.6$-fold. The direction is the same in all twelve cases: as a reference comes to resemble something the system saw in training, BLEU-4 accelerates and QA barely responds. 
    
        \begin{table*}[!t]
        \centering
        \small
        \setlength{\tabcolsep}{5pt}
        \renewcommand{\arraystretch}{1.15}
    
        \begin{tabular}{l cc cc c cc}
        \toprule
        & \multicolumn{2}{c}{\textbf{BLEU-1}}
        & \multicolumn{2}{c}{\textbf{BLEU-4}}
        & \textbf{QA \%}
        & \multicolumn{2}{c}{\textbf{Rank: BLEU-4 $\rightarrow$ QA}} \\
        \cmidrule(lr){2-3} \cmidrule(lr){4-5} \cmidrule(lr){6-6} \cmidrule(lr){7-8}
        \textbf{Model} & Reported & Ours & Reported & Ours & Ours & vs Reported & vs Ours \\
        \midrule
        \multicolumn{8}{l}{\textbf{PHOENIX-2014T}} \\
        \midrule
        C2RL         & 52.8 & 46.2 & 26.8 & $21.2_{\pm 2.2}$ & $52.6_{\pm 2.6}$ & 2 $\rightarrow$ 6 ($-4$) & 5 $\rightarrow$ 6 ($-1$) \\
        CiCo         &   -- & 47.6 &   -- & $23.6_{\pm 2.1}$ & $56.2_{\pm 2.6}$ & 3 $\rightarrow$ 2 ($+1$) & 1 $\rightarrow$ 2 ($-1$) \\
        Fla-LLM      & 46.3 & 46.6 & 23.1 & $21.1_{\pm 2.1}$ & $55.0_{\pm 2.5}$ & 4 $\rightarrow$ 3 ($+1$) & 6 $\rightarrow$ 3 ($+3$) \\
        GFSLT        & 43.7 & 46.9 & 21.4 & $22.2_{\pm 2.3}$ & $53.5_{\pm 2.6}$ & 6 $\rightarrow$ 5 ($+1$) & 4 $\rightarrow$ 5 ($-1$) \\
        SignCL       & 49.8 & 46.3 & 22.7 & $22.7_{\pm 2.2}$ & $53.8_{\pm 2.6}$ & 5 $\rightarrow$ 4 ($+1$) & 3 $\rightarrow$ 4 ($-1$) \\
        SingleStream & 54.0 & 46.0 & 27.6 & $23.5_{\pm 2.1}$ & $65.5_{\pm 2.3}$ & 1 $\rightarrow$ 1 ($\pm0$) & 2 $\rightarrow$ 1 ($+1$) \\
        \midrule
        \multicolumn{8}{l}{\textbf{CSL-Daily}} \\
        \midrule
        C2RL         & 49.3 & 30.6 & 21.6 & $\phantom{0}7.0_{\pm 0.7}$ & $11.6_{\pm 1.3}$ & 2 $\rightarrow$ 6 ($-4$) & 6 $\rightarrow$ 6 ($\pm0$) \\
        CiCo         &   -- & 39.9 &   -- & $12.0_{\pm 0.8}$ & $27.1_{\pm 1.8}$ & 5 $\rightarrow$ 3 ($+2$) & 3 $\rightarrow$ 3 ($\pm0$) \\
        Fla-LLM      & 37.1 & 47.8 & 14.2 & $16.9_{\pm 1.0}$ & $41.2_{\pm 2.1}$ & 4 $\rightarrow$ 2 ($+2$) & 2 $\rightarrow$ 2 ($\pm0$) \\
        GFSLT        & 39.4 & 37.4 & 11.0 & $10.9_{\pm 0.9}$ & $22.7_{\pm 1.7}$ & 6 $\rightarrow$ 4 ($+2$) & 4 $\rightarrow$ 4 ($\pm0$) \\
        SignCL       & 47.5 & 37.3 & 16.2 & $10.9_{\pm 0.8}$ & $21.2_{\pm 1.7}$ & 3 $\rightarrow$ 5 ($-2$) & 5 $\rightarrow$ 5 ($\pm0$) \\
        SingleStream & 55.4 & 58.8 & 25.8 & $28.2_{\pm 1.4}$ & $61.6_{\pm 2.1}$ & 1 $\rightarrow$ 1 ($\pm0$) & 1 $\rightarrow$ 1 ($\pm0$) \\
        \bottomrule
        \end{tabular}

        \caption{Reported vs.\ reproduced performance with QA
        content preservation. 95\% intervals are
        bootstrap over test instances (Appendix~\ref{app:ci}). QA is the pooled
        fraction of admitted questions answered correctly, from a bank
        generated and answered by \texttt{qwen2.5-32b}. Rank columns order all six models by
        BLEU-4 vs.\ QA: \emph{vs Reported} uses published BLEU-4 where available
        (our reproduction otherwise); \emph{vs Ours} uses our reproduction
        throughout.}
        \label{tab:combined_results}
        \end{table*}

              \paragraph{Probing sign grounding}
        \label{sec:grounding}

        One further perspective on the second question is worth detailing before we
        conclude. Figure~\ref{fig:input_modification} showed BLEU-4 surviving well above
        each model's noise floor under severe corruption of the input, but it cannot say
        what survived. A retention percentage reflects content the corruption removed,
        word order it broke, and target-side structure it never touched, and the two
        findings above say the third term dominates. What best withstands losing the
        visual signal is the same narrow, closed vocabulary that carries a third of the
        score and that training-like references inflate. Any metric resting on that
        scaffolding is inherently limited as a reflection of sign comprehension.

       Content words are harder to guess given their long-tailed distribution, but the share that survives pure noise shows they remain partly recoverable. A content-oriented metric is therefore not a fully closed channel, but a more transparent one. By stripping away the structural component that each modality expresses differently, the source of the non-isomorphism discussed in Section~\ref{sec:intro}, its movement under corruption can be read as grounding. What survives then reflects the content transferred without the masked channel. Neither function-word scaffolding nor train-test overlap inflates QA, whose slope in training-likeness stays flat where BLEU-4 accelerates (Figure~\ref{fig:sensitivity}), so its retention can be read directly: masking the hands leaves $64$--$80$\% of content transfer intact on Phoenix-2014T but only $4$--$16$\% on CSL-Daily, while masking the face leaves $62$--$70$\% and $40$--$68$\% respectively (Appendix~\ref{app:grounding}).

        In answer to the second evaluation question, content-based evaluation reduces that risk because it does not reward target-side regularities the way BLEU-4 does, turning a corruption experiment from an uninterpretable number into a per-channel statement about how far a system's output depends on the signing.

        \subsection{Saliency Oriented System Ranking}
        \label{sec:ranking}

        The third question asks how effectively models preserve the source discourse's
        salient content. QA accuracy ranges from $52.6$ to $65.5\%$ on Phoenix-2014T and
        from $11.6$ to $61.6\%$ on CSL-Daily (Table~\ref{tab:combined_results}). However,
        CSL-Daily's figures reflect total failures as much as partial transfer
        (Appendix~\ref{app:perinstance}). Relative to the published scores the ranking
        reorders substantially, but some of that is attributable to the reproduction gap, which is itself informative. The five gloss-free systems are evaluated here from
        the checkpoints of \citet{Sincan2025}, which brings them onto a common data
        pipeline, input resolution and evaluation protocol, and all six are decoded and
        scored through a single pipeline of our own (Appendix~\ref{app:datasets}). The
        \emph{Reported} column instead aggregates six independent training and evaluation
        setups, so we take the \emph{Ours} ranking to be the like-for-like comparison. On identical predictions the two metrics agree closely, no system changes rank on CSL-Daily, and five of six move by at most one position on Phoenix-2014T, since transferring more content also reproduces more reference surface. The exception is Fla-LLM, sixth on BLEU-4 and third on QA.

  The consequential finding is separation, not reranking. On Phoenix-2014T the five gloss-free systems cannot be told apart: all ten confidence-interval pairs overlap across a span of just $3.6$ QA points, so none shows evidence of preserving more salient content than another, on the field's most reported benchmark. CSL-Daily separates the same five over a span eight times wider. SingleStream, the one gloss-supervised system, stands apart from all five on both benchmarks ($9.3$ and $20.4$ points), a gap invisible to BLEU-4, which ranks it \emph{second} on Phoenix-2014T. In answer to the third evaluation question, preservation is partial at best: about two thirds of queried content for the gloss-supervised system on Phoenix-2014T, just above half for the gloss-free systems, and as little as a tenth on CSL-Daily.

    \section{Conclusion}

        BLEU-4 rewards words a spoken-language decoder can supply without reading the
        signing: it survives occlusion of the articulators, commits a third of its credit
        to function words with few counterparts in the source, and favours systems that
        reproduce training-like targets. Each exceeds the $2.5$-point spread separating the
        six systems we reproduce, so BLEU-4 differences at the current state of the art are
        not evidence of better sign understanding.

        Asking instead what a translation conveys answers all three questions of
        Section~\ref{sec:problem}. The QA pipeline measures salient content preservation
        reliably and at scale, stable to within half a point across reevaluations at
        $2$--$3$ GPU-minutes per additional system. It resists what inflates BLEU-4:
        $95\%$ of its credit falls on content, it is markedly more paraphrase-invariant and
        inversion-sensitive, and on the least training-like tenth of Phoenix-2014T it
        retains $57\%$ of its full-set value where BLEU-4 retains $12\%$. Measured this way
        the field looks different: models recover just above half the queried content, the
        gloss-supervised system leads by $9.3$ QA points where BLEU-4 ranks it second, and
        the five gloss-free systems cannot be told apart.
        
        As SLT matures, content transfer will increasingly saturate and be replaced by more
        detailed benchmarks. Until then, 
        we believe the field should focus on whether models are capable of recovering the
        salient, grounded content of the discourse.

\clearpage

{
    \small
    \bibliographystyle{ieeenat_fullname}
    \bibliography{main}

@inproceedings{zhang2019-paws,
    title = "{PAWS}: Paraphrase Adversaries from Word Scrambling",
    author = "Zhang, Yuan  and
      Baldridge, Jason  and
      He, Luheng",
    booktitle = "Proceedings of the 2019 Conference of the North {A}merican Chapter of the Association for Computational Linguistics: Human Language Technologies, Volume 1 (Long and Short Papers)",
    year = "2019"}

@article{sincan2025,
title = {Gloss-free Sign Language Translation: An unbiased evaluation of progress in the field},
journal = {Computer Vision and Image Understanding},
year = {2025},
author = {Ozge Mercanoglu Sincan and Jian He Low and Sobhan Asasi and Richard Bowden}
}

@article{raude2024,
    title={A Tale of Two Languages: Large-Vocabulary Continuous Sign Language
           Recognition from Spoken Language Supervision},
    author={Raude, Charles and Prajwal, K R and Momeni, Liliane and
            Bull, Hannah and Albanie, Samuel and
            Zisserman, Andrew and Varol, G{\"u}l},
    journal={arXiv},
    year={2024}
}

@InProceedings{Zuo2022,
    author    = {Zuo, Ronglai and Mak, Brian},
    title     = {C2SLR: Consistency-Enhanced Continuous Sign Language Recognition},
    booktitle = {Proceedings of the IEEE/CVF Conference on Computer Vision and Pattern Recognition (CVPR)},
    year      = {2022}
}

@inproceedings{brown2026,
  title     = {SignGPT and the Visual Language Toolkit},
  author    = {Brown, Matt and Ranum, Oline and Fish, Edward and Proctor, Heidi and Woll, Bencie and Bowden, Richard and Cormier, Kearsy},
  booktitle = {Proceedings of the LREC 2026 12th Workshop on the Representation and Processing of Sign Languages: Language in Motion},
  year      = {2026},
}

@inproceedings{kim2024,
    title = "{S}ign{BLEU}: Automatic Evaluation of Multi-channel Sign Language Translation",
    author = "Kim, Jung-Ho  and
      Huerta-Enochian, Mathew  and
      Ko, Changyong  and
      Lee, Du Hui",
    booktitle = "Proceedings of the 2024 Joint International Conference on Computational Linguistics, Language Resources and Evaluation (LREC-COLING 2024)",
    year = "2024",
    pages = "14796-14811",
}

@article{hamidullah2026,
      title={Grounding or Guessing? Visual Signals for Detecting Hallucinations in Sign Language Translation}, 
      author={Yasser Hamidullah and Koel Dutta Chowdhury and Yusser Al Ghussin and Shakib Yazdani and Cennet Oguz and Josef van Genabith and Cristina España-Bonet},
      year={2026},
      journal={arXiv} 
}

@inproceedings{shi2022,
    title = "Open-Domain Sign Language Translation Learned from Online Video",
    author = "Shi, Bowen  and
      Brentari, Diane  and
      Shakhnarovich, Gregory  and
      Livescu, Karen",
    booktitle = "Proceedings of the 2022 Conference on Empirical Methods in Natural Language Processing",
    year = "2022",

}

@inproceedings{
tanzer2025,
title={YouTube-{SL}-25: A Large-Scale, Open-Domain Multilingual Sign Language Parallel Corpus},
author={Garrett Tanzer and Biao Zhang},
booktitle={The Thirteenth International Conference on Learning Representations},
year={2025}
}

@INPROCEEDINGS {Yiting2023,
author = { Cheng, Yiting and Wei, Fangyun and Bao, Jianmin and Chen, Dong and Zhang, Wenqiang },
booktitle = { 2023 IEEE/CVF Conference on Computer Vision and Pattern Recognition (CVPR) },
title = { CiCo: Domain-Aware Sign Language Retrieval via Cross-Lingual Contrastive Learning },
year = {2023}}

@article{Chen2024a,
  title={C2RL: Content and Context Representation Learning for Gloss-Free Sign Language Translation and Retrieval},
  author={Zhigang Chen and Benjia Zhou and Yiqing Huang and Jun Wan and Yibo Hu and Hailin Shi and Yanyan Liang and Zhen Lei and Du Zhang},
  journal={IEEE Transactions on Circuits and Systems for Video Technology},
  year={2024}
}

@inproceedings{
fernandes2025,
title={Do {LLM}s Understand Your Translations? Evaluating Paragraph-level {MT} with Question Answering},
author={Patrick Fernandes and Sweta Agrawal and Emmanouil Zaranis and Andre Martins and Graham Neubig},
booktitle={Second Conference on Language Modeling},
year={2025}}

@inproceedings{zhang2025,
    title = "{L}i{T}rans{P}ro{QA}: An {LLM}-based Literary Translation Evaluation Metric with Professional Question Answering",
    author = "Zhang, Ran  and
      Zhao, Wei  and
      Macken, Lieve  and
      Eger, Steffen",
    booktitle = "Proceedings of the 2025 Conference on Empirical Methods in Natural Language Processing",
    year = "2025"
}

@inproceedings{krubinski2021,
    title = "Just Ask! Evaluating Machine Translation by Asking and Answering Questions",
    author = "Krubi{\'n}ski, Mateusz  and
      Ghadery, Erfan  and
      Moens, Marie-Francine  and
      Pecina, Pavel",
    booktitle = "Proceedings of the Sixth Conference on Machine Translation",
    year = "2021"
}

@inproceedings{han2022,
    title = "{S}im{QA}: Detecting Simultaneous {MT} Errors through Word-by-Word Question Answering",
    author = "Han, HyoJung  and
      Carpuat, Marine  and
      Boyd-Graber, Jordan",
    booktitle = "Proceedings of the 2022 Conference on Empirical Methods in Natural Language Processing",
    year = "2022"
}

@inproceedings{ki2025,
    title = "{A}sk{QE}: Question Answering as Automatic Evaluation for Machine Translation",
    author = "Ki, Dayeon  and
      Duh, Kevin  and
      Carpuat, Marine",
    booktitle = "Findings of the Association for Computational Linguistics: ACL 2025",
    year = "2025"}

@inproceedings{Kamalloo2024,
author = {Kamalloo, Ehsan and Upadhyay, Shivani and Lin, Jimmy},
title = {Towards Robust QA Evaluation via Open LLMs},
year = {2024},
booktitle = {Proceedings of the 47th International ACM SIGIR Conference on Research and Development in Information Retrieval}}

@inproceedings{chen2022,
author = {Chen, Yutong and Zuo, Ronglai and Wei, Fangyun and Wu, Yu and Liu, Shujie and Mak, Brian},
title = {Two-stream network for sign language recognition and translation},
year = {2022},
booktitle = {Proceedings of the 36th International Conference on Neural Information Processing Systems}}

@inproceedings{chen2024,
    title = "Factorized Learning Assisted with Large Language Model for Gloss-free Sign Language Translation",
    author = "Chen, Zhigang  and
      Zhou, Benjia  and
      Li, Jun  and
      Wan, Jun  and
      Lei, Zhen  and
      Jiang, Ning  and
      Lu, Quan  and
      Zhao, Guoqing",
    booktitle = "Proceedings of the 2024 Joint International Conference on Computational Linguistics, Language Resources and Evaluation (LREC-COLING 2024)",
    year = "2024"
}

@INPROCEEDINGS {zhou2023,
author = { Zhou, Benjia and Chen, Zhigang and Clapes, Albert and Wan, Jun and Liang, Yanyan and Escalera, Sergio and Lei, Zhen and Zhang, Du },
booktitle = { 2023 IEEE/CVF International Conference on Computer Vision (ICCV) },
title = {{ Gloss-free Sign Language Translation: Improving from Visual-Language Pretraining }},
year = {2023}}

@inproceedings{uthus2023,
author = {Uthus, David and Tanzer, Garrett and Georg, Manfred},
title = {YouTube-ASL: a large-scale, open-domain american sign language-english parallel corpus},
year = {2023},
booktitle = {Proceedings of the 37th International Conference on Neural Information Processing Systems}
}

@InProceedings{Albanie2021,
    author       = "Samuel Albanie and G{\"u}l Varol and Liliane Momeni and Hannah Bull and Triantafyllos Afouras and Himel Chowdhury and Neil Fox and Bencie Woll and Rob Cooper and Andrew McParland and Andrew Zisserman",
    title        = "{BOBSL}: {BBC}-{O}xford {B}ritish {S}ign {L}anguage {D}ataset",
    booktitle = {arXiv},
    year         = "2021",
}

@inproceedings{Duarte2021,
    title={{How2Sign: A Large-scale Multimodal Dataset for Continuous American Sign Language}},
    author={Duarte, Amanda and Palaskar, Shruti and Ventura, Lucas and Ghadiyaram, Deepti and DeHaan, Kenneth and
                   Metze, Florian and Torres, Jordi and Giro-i-Nieto, Xavier},
    booktitle={Conference on Computer Vision and Pattern Recognition (CVPR)},
    year={2021}
}

@INPROCEEDINGS {Zhou2021,
author = { Zhou, Hao and Zhou, Wengang and Qi, Weizhen and Pu, Junfu and Li, Houqiang },
booktitle = { 2021 IEEE/CVF Conference on Computer Vision and Pattern Recognition (CVPR) },
title = {{ Improving Sign Language Translation with Monolingual Data by Sign Back-Translation }},
year = {2021}}

@ARTICLE{alakin2026,
    
AUTHOR={Alkain, Bittor  and Núñez-Marcos, Adrián  and Escolano, Carlos  and Docío-Fernández, Laura  and Perez-de-Viñaspre, Olatz  and Labaka, Gorka },
           
TITLE={Critical analysis of datasets for sign language translation},
          
JOURNAL={Frontiers in Artificial Intelligence},
  
YEAR={2026}
}

@inproceedings{Jinhui2024,
author = {Ye, Jinhui and Wang, Xing and Jiao, Wenxiang and Liang, Junwei and Xiong, Hui},
title = {Improving gloss-free sign language translation by reducing representation density},
year = {2024},
booktitle = {Proceedings of the 38th International Conference on Neural Information Processing Systems}
}

@InProceedings{Camgoz_2018_CVPR,
author = {Camgoz, Necati Cihan and Hadfield, Simon and Koller, Oscar and Ney, Hermann and Bowden, Richard},
title = {Neural Sign Language Translation},
booktitle = {Proceedings of the IEEE Conference on Computer Vision and Pattern Recognition (CVPR)},
year = {2018}
}

@misc{fashn-human-parser,
    author = {AI FASHN},
  title = {FASHN Human Parser: SegFormer for Fashion Human Parsing},
  year = {2024},
  publisher = {Hugging Face},
  url = {https://huggingface.co/fashn-ai/fashn-human-parser}
}

@inproceedings{bragg2019sign,
  title={Sign language recognition, generation, and translation: An interdisciplinary perspective},
  author={Bragg, Danielle and Koller, Oscar and Bellard, Marc and others},
  booktitle={ASSETS},
  year={2019}
}

@article{desai2024systemic,
  title={Systemic Biases in Sign Language AI Research: A Deaf-Led Call to Reevaluate Research Agendas},
  author={Desai, A. and De Meulder, M. and Hochgesang, J. A. and Kocab, A. and Lu, A. X.},
  journal={LREC-COLING Workshop},
  year={2024}
}

@inproceedings{fox2023best,
  title={Best practices for sign language technology research},
  author={Fox, N. and Woll, B. and Cormier, K.},
  booktitle={Universal Access in the Information Society},
  year={2023}
}

@inproceedings{chen2023twostream,
  title={Two-Stream Network for Sign Language Recognition and Translation},
  author={Chen, Y. and Zuo, R. and Wei, F. and others},
  booktitle={NeurIPS},
  year={2023}
}

@inproceedings{chen2023simple,
  title={A Simple Multi-Modality Transfer Learning Baseline for Sign Language Translation},
  author={Chen, Y. and Wei, F. and Sun, X. and others},
  booktitle={CVPR},
  year={2023}
}

@inproceedings{yin2021simulslt,
  title={SimulSLT: End-to-End Simultaneous Sign Language Translation},
  author={Yin, A. and Zhao, Z. and Liu, J. and others},
  booktitle={ACM MM},
  year={2021}
}

@inproceedings{zhou2021improving,
  title={Improving Sign Language Translation with Monolingual Data by Sign Back-Translation},
  author={Zhou, H. and Zhou, W. and Qi, W. and others},
  booktitle={CVPR},
  year={2021}
}

@article{zhou2022spatial,
  title={Spatial-Temporal Multi-Cue Network for Sign Language Recognition and Translation},
  author={Zhou, H. and Zhou, W. and Li, H.},
  journal={IEEE Transactions on Multimedia},
  year={2022}
}

@inproceedings{zhang2023sltunet,
  title={SLTUNET: A Simple Unified Model for Sign Language Translation},
  author={Zhang, B. and Müller, M. and Sennrich, R.},
  booktitle={ICLR},
  year={2023}
}

@inproceedings{wong2024sign2gpt,
  title={Sign2GPT: Leveraging Large Language Models for Gloss-Free Sign Language Translation},
  author={Wong, R. and Camgoz, N. C. and Bowden, R.},
  booktitle={ICLR},
  year={2024}
}

@inproceedings{kim2025leveraging,
  title={Leveraging the Power of MLLMs for Gloss-Free Sign Language Translation},
  author={Kim, J. and Jeon, H. and Bae, J. and Kim, H. Y.},
  booktitle={ICCV},
  year={2025}
}

@article{chen2025c2rl,
  title={C2RL: Content and Context Representation Learning for Gloss-Free Sign Language Translation and Retrieval},
  author={Chen, Z. and Zhou, B. and Huang, Y. and others},
  journal={IEEE TCSVT},
  year={2025}
}

@inproceedings{zhou2023glossfree,
  title={Gloss-free Sign Language Translation: Improving from Visual-Language Pretraining},
  author={Zhou, B. and Chen, Z. and Wan, J. and others},
  booktitle={ICCV},
  year={2023}
}

@inproceedings{akhbardeh2021,
    title = "Findings of the 2021 Conference on Machine Translation ({WMT}21)",
    author = "Akhbardeh, Farhad  and
      Arkhangorodsky, Arkady  and
      Biesialska, Magdalena  and
      Bojar, Ond{\v{r}}ej  and
      Chatterjee, Rajen  and
      Chaudhary, Vishrav  and
      Costa-jussa, Marta R.  and
      Espa{\~n}a-Bonet, Cristina  and
      Fan, Angela  and
      Federmann, Christian  and
      Freitag, Markus  and
      Graham, Yvette  and
      Grundkiewicz, Roman  and
      Haddow, Barry  and
      Harter, Leonie  and
      Heafield, Kenneth  and
      Homan, Christopher M.  and
      Huck, Matthias  and
      Amponsah-Kaakyire, Kwabena  and
      Kasai, Jungo  and
      Khashabi, Daniel  and
      Knight, Kevin  and
      Kocmi, Tom  and
      Koehn, Philipp  and
      Lourie, Nicholas  and
      Monz, Christof  and
      Morishita, Makoto  and
      Nagata, Masaaki  and
      Nagesh, Ajay  and
      Nakazawa, Toshiaki  and
      Negri, Matteo  and
      Pal, Santanu  and
      Tapo, Allahsera Auguste  and
      Turchi, Marco  and
      Vydrin, Valentin  and
      Zampieri, Marcos",
    booktitle = "Proceedings of the Sixth Conference on Machine Translation",
    year = "2021"
}

@inproceedings{asasi2025beyond,
  title={Beyond Gloss: A Hand-Centric Framework for Gloss-Free Sign Language Translation},
  author={Asasi, S. and Lakhal, M. I. and Sincan, O. M. and Bowden, R.},
  year={2025},
  booktitle={arXiv preprint}
}

@article{freitag2021,
    title = "Experts, Errors, and Context: A Large-Scale Study of Human Evaluation for Machine Translation",
    author = "Freitag, Markus  and
      Foster, George  and
      Grangier, David  and
      Ratnakar, Viresh  and
      Tan, Qijun  and
      Macherey, Wolfgang",
    journal = "Transactions of the Association for Computational Linguistics",
    year = "2021"
}

@inproceedings{callison2011,
    title = "Findings of the 2011 Workshop on Statistical Machine Translation",
    author = "Callison-Burch, Chris  and
      Koehn, Philipp  and
      Monz, Christof  and
      Zaidan, Omar",
    booktitle = "Proceedings of the Sixth Workshop on Statistical Machine Translation",
    year = "2011"
}

@inproceedings{callison2012,
    title = "Findings of the 2012 Workshop on Statistical Machine Translation",
    author = "Callison-Burch, Chris  and
      Koehn, Philipp  and
      Monz, Christof  and
      Post, Matt  and
      Soricut, Radu  and
      Specia, Lucia",
    booktitle = "Proceedings of the Seventh Workshop on Statistical Machine Translation",
    year = "2012"
}

@inproceedings{bojar2013,
    title = "Findings of the 2013 {W}orkshop on {S}tatistical {M}achine {T}ranslation",
    author = "Bojar, Ond{\v{r}}ej  and
      Buck, Christian  and
      Callison-Burch, Chris  and
      Federmann, Christian  and
      Haddow, Barry  and
      Koehn, Philipp  and
      Monz, Christof  and
      Post, Matt  and
      Soricut, Radu  and
      Specia, Lucia",
    booktitle = "Proceedings of the Eighth Workshop on Statistical Machine Translation",
    year = "2013"
}

@inproceedings{bojar2014,
    title = "Findings of the 2014 Workshop on Statistical Machine Translation",
    author = "Bojar, Ond{\v{r}}ej  and
      Buck, Christian  and
      Federmann, Christian  and
      Haddow, Barry  and
      Koehn, Philipp  and
      Leveling, Johannes  and
      Monz, Christof  and
      Pecina, Pavel  and
      Post, Matt  and
      Saint-Amand, Herve  and
      Soricut, Radu  and
      Specia, Lucia  and
      Tamchyna, Ale{\v{s}}",
    booktitle = "Proceedings of the Ninth Workshop on Statistical Machine Translation",
    year = "2014"
}

@inproceedings{bojar2015,
    title = "Findings of the 2015 Workshop on Statistical Machine Translation",
    author = "Bojar, Ond{\v{r}}ej  and
      Chatterjee, Rajen  and
      Federmann, Christian  and
      Haddow, Barry  and
      Huck, Matthias  and
      Hokamp, Chris  and
      Koehn, Philipp  and
      Logacheva, Varvara  and
      Monz, Christof  and
      Negri, Matteo  and
      Post, Matt  and
      Scarton, Carolina  and
      Specia, Lucia  and
      Turchi, Marco",
    booktitle = "Proceedings of the Tenth Workshop on Statistical Machine Translation",
    year = "2015"
}

@inproceedings{buckkoehn2016,
    title = "Findings of the {WMT} 2016 Bilingual Document Alignment Shared Task",
    author = "Buck, Christian  and
      Koehn, Philipp",
    booktitle = "Proceedings of the First Conference on Machine Translation: Volume 2, Shared Task Papers",
    year = "2016"
}

@inproceedings{bojar2017,
    title = "Findings of the 2017 Conference on Machine Translation ({WMT}17)",
    author = "Bojar, Ond{\v{r}}ej  and
      Chatterjee, Rajen  and
      Federmann, Christian  and
      Graham, Yvette  and
      Haddow, Barry  and
      Huang, Shujian  and
      Huck, Matthias  and
      Koehn, Philipp  and
      Liu, Qun  and
      Logacheva, Varvara  and
      Monz, Christof  and
      Negri, Matteo  and
      Post, Matt  and
      Rubino, Raphael  and
      Specia, Lucia  and
      Turchi, Marco",
    booktitle = "Proceedings of the Second Conference on Machine Translation",
    year = "2017"
}

@InProceedings{creutz2018,
  title = {Open Subtitles Paraphrase Corpus for Six Languages},
  author={Mathias Creutz},
  booktitle={Proceedings of the 11th edition of the Language Resources and Evaluation Conference (LREC 2018)},
  year={2018}
  }

@inproceedings{freitag2020,
    title = "Human-Paraphrased References Improve Neural Machine Translation",
    author = "Freitag, Markus  and
      Foster, George  and
      Grangier, David  and
      Cherry, Colin",
    booktitle = "Proceedings of the Fifth Conference on Machine Translation",
    year = "2020"
}

@inproceedings{jiang2025,
    title = "Meaningful Pose-Based Sign Language Evaluation",
    author = {Jiang, Zifan  and
      Leong, Colin  and
      Moryossef, Amit  and
      Cory, Oliver  and
      Ivashechkin, Maksym  and
      Tarigopula, Neha  and
      Zhang, Biao  and
      G{\"o}hring, Anne  and
      Rios, Annette  and
      Sennrich, Rico  and
      Ebling, Sarah},
    booktitle = "Proceedings of the Tenth Conference on Machine Translation",
    year = "2025"
}

@article{xia2026,
author = {Xia, Yibo and Zhan, Qihui and Luo, Xiaoyan and Shi, Xiaofeng and Wang, Yunhong},
title = {SignMask: Structure-aware Masked Modeling for Holistic 3D Sign Language Production},
year = {2026},
journal = {ACM Trans. Multimedia Comput. Commun. Appl.}
}

@inproceedings{
wang2025,
title={Advanced Sign Language Video Generation with Compressed and Quantized Multi-Condition Tokenization},
author={Cong Wang and Zexuan Deng and Zhiwei Jiang and Yafeng Yin and Fei Shen and Zifeng Cheng and Shiping Ge and Shiwei Gan and Qing Gu},
booktitle={The Thirty-ninth Annual Conference on Neural Information Processing Systems},
year={2025}}

@inproceedings{yin2024,
    title = "{T}2{S}-{GPT}: Dynamic Vector Quantization for Autoregressive Sign Language Production from Text",
    author = "Yin, Aoxiong  and
      Li, Haoyuan  and
      Shen, Kai  and
      Tang, Siliang  and
      Zhuang, Yueting",
    booktitle = "Proceedings of the 62nd Annual Meeting of the Association for Computational Linguistics (Volume 1: Long Papers)",
    year = "2024"}

@article{Xie2023,
  title={Sign Language Production with Latent Motion Transformer},
  author={Pan Xie and Taiying Peng and Yao Du and Qipeng Zhang},
  journal={2024 IEEE/CVF Winter Conference on Applications of Computer Vision (WACV)},
  year={2023}}

@inproceedings{Zhengdi2024,
author = {Yu, Zhengdi and Huang, Shaoli and Cheng, Yongkang and Birdal, Tolga},
title = {SignAvatars: A Large-Scale 3D Sign Language Holistic Motion Dataset and Benchmark},
year = {2024},
booktitle = {Computer Vision – ECCV 2024: 18th European Conference, Milan, Italy, September 29–October 4, 2024, Proceedings, Part V}}

@inproceedings{
ranum2024,
title={The {NGT}200 Dataset - Geometric Multi-View Isolated Sign Recognition},
author={Oline Ranum and David Wessels and Gom{\`e}r Otterspeer and Erik J Bekkers and Floris Roelofsen and Jari I. Andersen},
booktitle={ICML 2024 Workshop on Geometry-grounded Representation Learning and Generative Modeling},
year={2024}
}

@article{ranum2026,
      title={What's the Point? Spatial Grammar \& Index Resolution for Sign Language Processing}, 
      author={Oline Ranum and Simon Hadfield and Richard Bowden},
      year={2026},
      journal={arXiv}
}

@inproceedings{Huang2018,
author = {Huang, Jie and Zhou, Wengang and Zhang, Qilin and Li, Houqiang and Li, Weiping},
title = {Video-based sign language recognition without temporal segmentation},
year = {2018},
booktitle = {Proceedings of the Thirty-Second AAAI Conference on Artificial Intelligence and Thirtieth Innovative Applications of Artificial Intelligence Conference and Eighth AAAI Symposium on Educational Advances in Artificial Intelligence}
}

@article{RASTGOO2021,
title = {Sign Language Recognition: A Deep Survey},
journal = {Expert Systems with Applications},
year = {2021},
author = {Razieh Rastgoo and Kourosh Kiani and Sergio Escalera}}

@inproceedings{li2020,
    title={Word-level Deep Sign Language Recognition from Video: A New Large-scale Dataset and Methods Comparison},
    author={Li, Dongxu and Rodriguez, Cristian and Yu, Xin and Li, Hongdong},
    booktitle={The IEEE Winter Conference on Applications of Computer Vision},
   year={2020}
 }

@article{Saunders2020,
title = {Adversarial Training for Multi-Channel Sign Language Production},
author = {Saunders, Ben and Camgöz, Necati Cihan and Bowden, Richard},
journal = {The 31st British Machine Vision Virtual Conference},
year = {2020}
}

@inproceedings{yang2019,
    title = "{PAWS}-{X}: A Cross-lingual Adversarial Dataset for Paraphrase Identification",
    author = "Yang, Yinfei  and
      Zhang, Yuan  and
      Tar, Chris  and
      Baldridge, Jason",
    booktitle = "Proceedings of the 2019 Conference on Empirical Methods in Natural Language Processing and the 9th International Joint Conference on Natural Language Processing (EMNLP-IJCNLP)",
    year = "2019"
}

@article{cory2026,
  title   = {BackTranslation2.0: A Linguistically Motivated Metric
             to Assess Sign Language Production},
  author  = {Cory, Oliver and Ivashechkin, Maksym and Sahin, Karahan and
             Ranum, Oline and Low, Jianhe and Fish, Edward and
             Pelykh, Anton and Mercanoglu Sincan, Ozge and
             Bowden, Richard},
  journal = {arXiv preprint arXiv:2606.28673},
  year    = {2026},
}

@inproceedings{kwon2023,
  title={Efficient Memory Management for Large Language Model Serving with PagedAttention},
  author={Woosuk Kwon and Zhuohan Li and Siyuan Zhuang and Ying Sheng and Lianmin Zheng and Cody Hao Yu and Joseph E. Gonzalez and Hao Zhang and Ion Stoica},
  booktitle={Proceedings of the ACM SIGOPS 29th Symposium on Operating Systems Principles},
  year={2023}
}

@ARTICLE{Rosenburg2020,
    
AUTHOR={Rosenburg, Patrick  and Lieberman, Amy M.  and Caselli, Naomi  and Hoffmeister, Robert },
           
TITLE={The Development and Evaluation of a New ASL Text Comprehension Task},
          
JOURNAL={Frontiers in Communication},
    
  
YEAR={2020}}

@article{Hauser2016,
  title={American Sign Language Comprehension Test: A Tool for Sign Language Researchers.},
  author={Peter C. Hauser and Raylene Paludneviciene and Wanda Riddle and Kim Brown Kurz and Karen Emmorey and Jessica Contreras},
  journal={Journal of deaf studies and deaf education},
  year={2016}}

@article{papineni2002bleu,
  title={BLEU: a Method for Automatic Evaluation of Machine Translation},
  author={Papineni, K. and Roukos, S. and Ward, T. and Zhu, W. J.},
  journal={ACL},
  year={2002}
}
}

\clearpage

\section{Supplementary Material}

        \appendix

        \setcounter{topnumber}{3}
        \setcounter{bottomnumber}{1}
        \setcounter{totalnumber}{4}
        \renewcommand{\topfraction}{0.92}
        \renewcommand{\textfraction}{0.06}
        \renewcommand{\floatpagefraction}{0.80}
        \renewcommand{\dbltopfraction}{0.92}
        \renewcommand{\dblfloatpagefraction}{0.70}


    \section{Datasets and Reproduced Systems}
        \label{app:datasets}
    
        \subsection{Spoken-language benchmarks}
    
        \paragraph{WMT multi-reference collections.} The WMT datasets are the parallel
        corpora and human judgements released with the annual Conference on Machine
        Translation shared tasks, each denoted WMT\textit{xx} for the year of the task.
        Most releases provide a single reference per source segment and are therefore
        unusable for paraphrase invariance, which needs several equally valid renderings
        of the same content.
    
        \emph{WMT19-paraphrased} \cite{freitag2020} augments the 1997 newstest2019
        en--de segments with six line-aligned human reference sets produced
        under different paraphrasing instructions and quality-control regimes. Every set
        is a faithful rendering of the same source, so content is constant by
        construction and only wording varies. After deduplicating identical renderings
        all 1997 segments are retained, averaging $3.46$ distinct variants.
    
        \emph{WMT21-multiref} uses the newstest2021 en--de references A, C and D, three
        independently produced human translations of each segment, and recovers the
        ref-B text from the raw MQM judgement file for segments that were rated. Keeping
        segments with at least three distinct variants yields 991 groups at $3.42$
        variants each.
    
        Within each group one rendering is designated the reference (by default the variant with the highest mean BLEU to the others); the
        remaining in-group renderings supply the paraphrase scores and two randomly
        drawn out-of-group renderings supply the floor
        (Appendix~\ref{app:metricdefs}).
    
        \paragraph{WMT human concordance.} The system-ranking analysis of
        Section~\ref{sec:qatext} uses the human judgements released with the campaigns
        from 2011 to 2024, under three successive label regimes: relative ranking (RR,
        2011--2016), DA-derived relative ranking (daRR, 2017--2019), and MQM
        (2020--2024), from which pairwise preferences are derived at a severity margin
        of $1.0$. Metric-tuned MBR submissions are excluded at every call site.
    
        \paragraph{OpusParcus.} The OpenSubtitles Paraphrase Corpus \cite{creutz2018}
        covers six European languages (en, de, fi, fr, ru, sv). Extracted from
        OpenSubtitles2016 film and television subtitles, it reflects informal,
        colloquial and often elliptical language, and its sentences are short, so a
        metric has few tokens to work with. Its development and test sets carry a graded
        human annotation score from 1 (not a paraphrase) to 4 (clean paraphrase),
        obtained by averaging two independent annotators; the label is therefore
        half-integer valued and encodes annotator disagreement rather than discarding
        it. We pool the validation and test splits and evaluate all 20{,}378 pairs. The
        binned analysis groups the scale into Q1 $[1,1.5]$, Q2 $[2,2.5]$, Q3 $[3,3.5]$
        and Q4 $=4.0$.
    
        \paragraph{PAWS-X.} PAWS-X \cite{yang2019} is the cross-lingual extension of
        PAWS \cite{zhang2019-paws} (Paraphrase Adversaries from Word Scrambling),
        covering seven languages (en, de, fr, es, zh, ja, ko) with 2{,}000 test pairs
        each. Pairs are adversarially constructed so surface overlap is high in both
        classes, with negatives generated by word scrambling and argument swaps. We use
        the test split and the binary label (1 = true paraphrase, 0 = meaning-flipped)
        as ground truth. 
    
        \subsection{Sign language benchmarks}
    
        \paragraph{Phoenix-2014T} \cite{Camgoz_2018_CVPR} covers
        German Sign Language weather forecasting, interpreted for the PHOENIX broadcaster
        by nine signers. Its $8{,}257$ segments split $7{,}096 / 519 / 642$ across train,
        dev and test. The test split carries $7{,}816$ German tokens over $1{,}001$ word
        types, averaging $12.2$ tokens per sentence, against $4{,}264$ gloss tokens over
        $411$ types. The vocabulary is narrow and nearly closed with respect to training:
        the $2{,}887$ training word types cover all but $59$ test types, a token OOV rate
        of $0.77\%$. References are distributed lowercased and contain no punctuation or
        digits, and $630$ of the $642$ are distinct.

        \paragraph{CSL-Daily} \cite{Zhou2021} covers Chinese Sign Language on daily-life
        topics, signed by ten signers, with $20{,}654$ segments split
        $18{,}401 / 1{,}077 / 1{,}176$ (one training row carries an empty target and is
        dropped). Chinese is written unspaced, so counts are per character: the test
        split has $19{,}239$ characters over $1{,}346$ types, $16.4$ per sentence,
        against $9{,}002$ gloss tokens over $1{,}345$ types drawn from a training gloss
        vocabulary of exactly $2{,}000$. Character OOV is $0.31\%$. Every reference
        retains full-width punctuation, and because each phrase is signed by several
        signers the split holds only $798$ distinct sentences among its $1{,}176$
        references.

                \paragraph{Reference preprocessing.} BLEU-4 is corpus BLEU-4 under the tokeniser
        policy of Appendix~\ref{app:metricdefs}, so Phoenix-2014T references are scored
        exactly as distributed. On CSL-Daily the systems emit character-spaced Chinese
        while the references are unspaced, which drives BLEU-4 to zero if left
        uncorrected; we therefore strip whitespace adjacent to any CJK character and fold
        full-width punctuation to ASCII in both hypothesis and reference before scoring.
        The QA protocol reads the same references with no further normalisation.

        \paragraph{Reproduced models.} The six systems of
        Section~\ref{sec:bleulimits}, condensed from Section~2 for space.
        GFSLT-VLP \cite{zhou2023} pretrains a visual encoder and text decoder via
        contrastive vision-language alignment and masked language modeling, then
        fine-tunes an encoder-decoder for translation.
        FLA-LLM \cite{chen2024} pretrains its visual encoder with a lightweight
        translation decoder, then freezes it and pairs it with a larger 12-layer MBart.
        CiCo \cite{Yiting2023} performs sign-video-text retrieval via cross-lingual
        contrastive learning, capturing fine-grained sign-to-word mappings in a joint
        embedding space.
        SignCL \cite{Jinhui2024} applies a contrastive loss between adjacent frames to
        counter representation density and better separate distinct signs.
        C2RL \cite{Chen2024a} combines visual-text alignment with a lightweight
        translation objective in a multi-task pretraining framework.
        TwoStream \cite{chen2022} combines RGB video and keypoint sequences through
        lateral connections, a sign pyramid network, and self-distillation to reduce
        visual redundancy; this work uses only its single-stream (RGB-only) variant, the
        one gloss-supervised system in the set.
    
         \paragraph{Checkpoints.} Every system is reproduced from released weights rather
        than retrained, so Table~\ref{tab:combined_results} reflects the published models
        and not our own training runs. The five gloss-free systems use the checkpoints
        distributed with the SLTBaselines benchmark of \citet{Sincan2025}, which
        re-evaluates GFSLT-VLP, FLA-LLM, CiCo, SignCL and C2RL under a common data
        pipeline, input resolution and evaluation protocol. We take one checkpoint per
        model and dataset, together with the trimmed mBART tokeniser and visual-encoder
        configuration shipped in the same bundle, since the released weights are
        vocabulary-consistent only with that pair. The single-stream variant of TwoStream
        \cite{chen2022} is not part of that benchmark and comes from the authors' own
        TwoStreamNetwork release. Drawing the gloss-free systems from one reproduction
        source is what makes them comparable here: their originally reported scores were
        obtained under six different training and evaluation setups, which is the
        discrepancy \citet{Sincan2025} set out to quantify and which the gap between the
        \emph{Reported} and \emph{Ours} columns of Table~\ref{tab:combined_results}
        reflects.

        \section{Input Corruption: Protocol and Retention}
        \label{app:inputcorruption}

        All corruptions are applied at inference time to the reproduced checkpoints.
        Each condition is decoded over the full test split of both datasets, and
        retention is reported as a percentage of that model's own undistorted baseline,
        so architectures with different absolute scores remain comparable. The BLEU-4
        retentions are Figure~\ref{fig:input_modification}; the QA retentions on the
        same conditions are given in Section~\ref{app:grounding} below.

        \subsection{Temporal frame shuffling}

        Each clip's frames are randomly permuted before the visual encoder, with a
        permutation seeded per sample to ensure reproducibility. The permutation is
        length-preserving and leaves the padding frames the dataloader adds in place, so
        every downstream sequence length and attention mask is unchanged.

        \begin{figure}[!t]
            \centering
            \includegraphics[width=\linewidth]{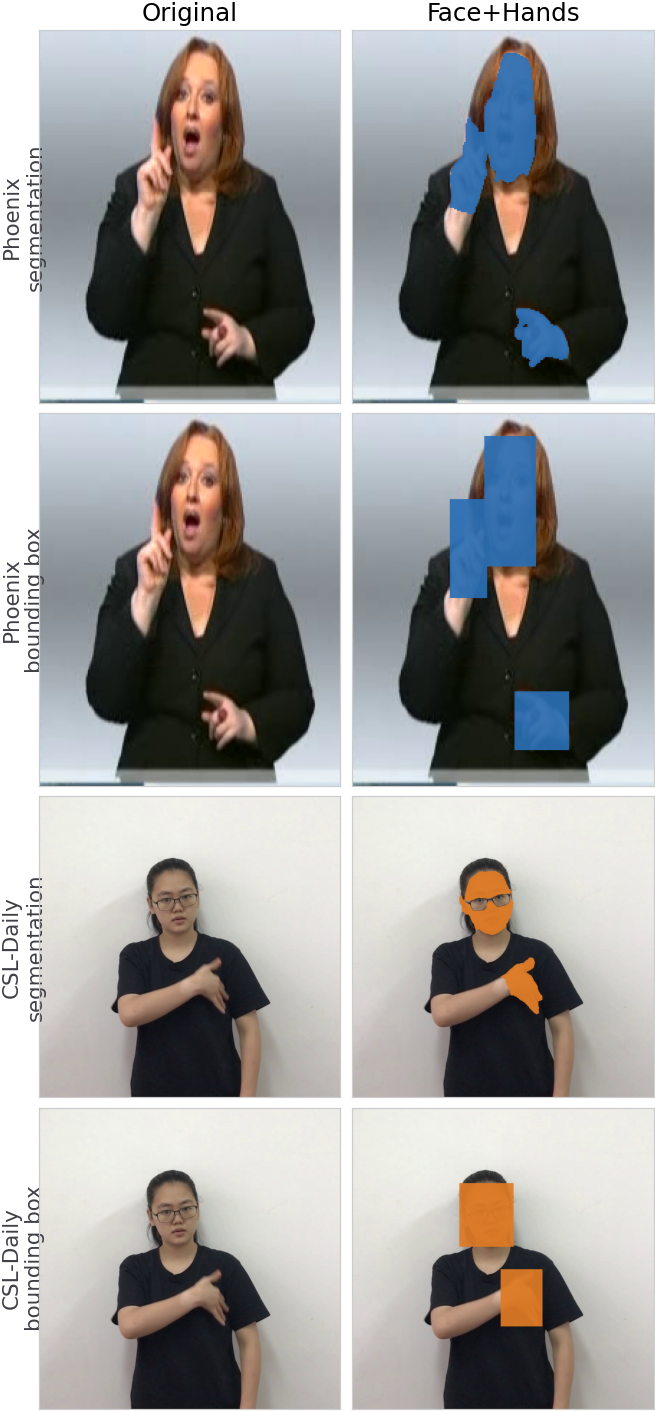}
            \caption{The two spatial-mask variants, on one Phoenix-2014T frame (blue)
            and one CSL-Daily frame (orange). \emph{Segmentation} rows occlude the
            region traced by the segmentation model and are used for the main results;
            \emph{bounding box} rows occlude each connected component's filled bounding
            box. The occlusion is drawn in each dataset's identity colour rather than
            black, which would read as part of the signer's dark clothing. Note that the
            segmentation masks preserve the outline of the occluded articulator.}
            \label{fig:maskvariants}
        \end{figure}

        \subsection{Spatial masking}

\begin{figure}[!h]
            \centering
            \includegraphics[width=\linewidth]{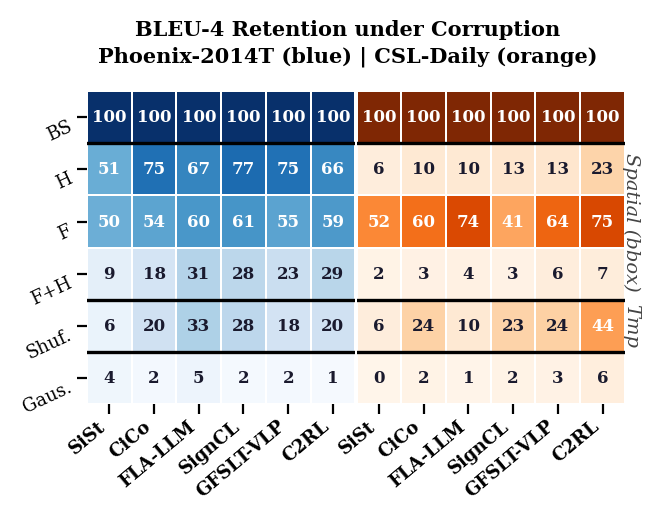}\\[0.6em]
            \includegraphics[width=\linewidth]{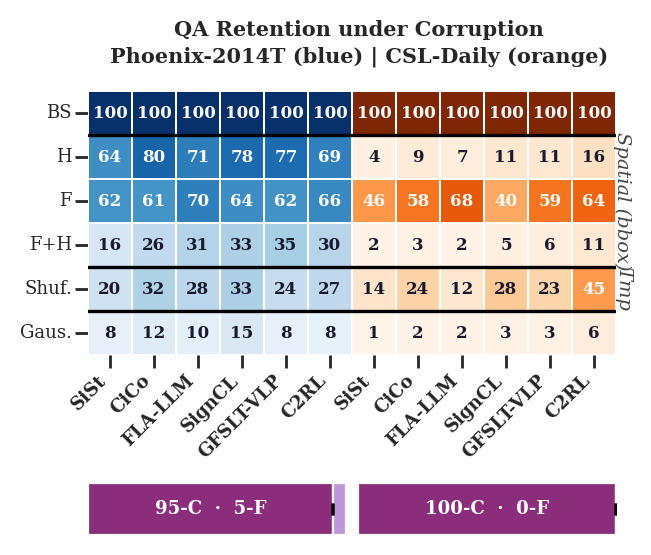}
            \caption{Retention under the bounding-box mask variant, BLEU-4 (top) and QA
            (bottom), to be read against Figures~\ref{fig:input_modification}
            and~\ref{fig:qaretheat}. Boxing the articulator is the harsher occlusion and
            costs roughly ten further points of retention throughout, but the ordering of
            conditions and the differing hand/face profiles of the two datasets are
            unchanged, so no conclusion in the main text depends on the choice of mask.}
            \label{fig:maskvariantsretentions}
        \end{figure}

        The masking ablations occlude the pixels of a body-part region in every frame
        before the visual encoder, using precomputed per-video masks for the face and the
        hands. The masks are produced by a publicly available segmentation model
        \cite{fashn-human-parser} that traces the outline of each articulator. The
        resulting mask retains the silhouette of what it removes, so a hand-shaped void
        may still expose the handshape. We therefore repeat every masking condition with
        a bounding-box variant, in which each connected component of a region is replaced
        by its filled axis-aligned box. Boxing per component rather than per region
        matters: the two hands are separate components, and a single box spanning both
        would occlude the torso between them. Figure~\ref{fig:maskvariants} shows both
        variants on both datasets, and Figure~\ref{fig:maskvariantsretentions} reports
        the corresponding retention scores.

        \subsection{Gaussian noise}

        This condition replaces the visual representation outright, giving a floor in
        which no visual evidence survives, so whatever a model still scores is
        attributable to target-side priors alone. Where the replacement is applied
        differs between the two model families, because they expose different
        interception points, and the difference is worth stating rather than eliding.

        The five gloss-free systems are corrupted at the encoder output. The visual
        encoder runs untouched and its $T \times D$ feature sequence $E$ is then
        overwritten elementwise by draws from $\mathcal{N}(\hat\mu_E, \hat\sigma_E^2)$,
        where $\hat\mu_E$ and $\hat\sigma_E$ are the mean and standard deviation of $E$
        itself, so the decoder is handed a sequence of the original length and scale that
        carries no information. The attention mask is passed through unchanged, so padded
        positions stay masked and every downstream length is as at baseline. SingleStream
        exposes no equivalent point, so there the raw frame tensor is overwritten instead:
        every pixel of every frame and channel is redrawn from a Gaussian matched to the
        clip's own mean and standard deviation, clamped back into the valid pixel range,
        and passed through the real encoder. Both paths compute their statistics per clip,
        the ablation decoding running at batch size one.

        The two are therefore not the identical operation, one destroying the encoded
        representation and the other the input to the encoder, and a reader should not
        treat the noise row as a single controlled manipulation across all six systems.
        What licenses reading it as one floor is that the two cut points are empirically
        interchangeable at this severity: BLEU-4 retention under noise is $1$--$5\%$
        across the five feature-level models on Phoenix-2014T and $1$--$6\%$ on
        CSL-Daily, and SingleStream, the pixel-level one, retains $4.0\%$ and $0.4\%$,
        inside that range on Phoenix-2014T and just below it on CSL-Daily
        (Figure~\ref{fig:input_modification}). Neither cut leaves a model anything to
        read.

        \subsection{Retention under both metrics}
        \label{app:grounding}

\begin{figure}[!h]
            \centering
            \includegraphics[width=\linewidth]{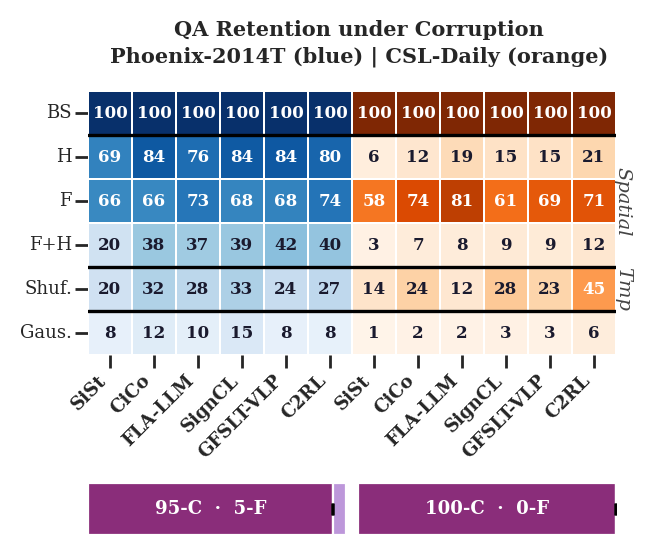}
            \caption{QA accuracy retention (\% of baseline) under the same input
            corruptions as Figure~\ref{fig:input_modification}, over all six models.
            Read down a column for one model's dependence on each channel. The
            bounding-box mask variant is in Figure~\ref{fig:maskvariantsretentions}.}
            \label{fig:qaretheat}
        \end{figure}

        Section~\ref{sec:grounding} states the headline profiles; this section gives the
        figure and the remaining conditions. Figure~\ref{fig:qaretheat} repeats the
        corruption experiment of Figure~\ref{fig:input_modification} under QA. On
        Phoenix-2014T, occluding both hands still leaves $64$--$80\%$ of content transfer
        intact and occluding the face leaves $62$--$70\%$; only removing both together
        brings models down to $16$--$35\%$. On CSL-Daily the hands retain just
        $4$--$16\%$ while the face retains $40$--$68\%$. Neither dataset's models depend
        on temporal order as much as on either articulator: shuffling frames retains
        $20$--$33\%$ and $12$--$45\%$. Replacing the signal with noise floors both, at
        $8$--$15\%$ and $1$--$6\%$, the residue attributable to target-side priors alone.

        Two readings follow. Models trained on Phoenix-2014T transfer most of their
        measured content without the manual channel, which is difficult to reconcile with
        predictions grounded in the signing, and the dataset's narrow weather-forecast
        scope is the obvious suspect. And because the two corpora reverse which
        articulator is load-bearing under fixed architectures, grounding is a property of
        what the benchmark affords. Neither reading would be legible from an n-gram
        score.

        \section{Content and Function Words: Tagging, Distribution and Attribution}
        \label{app:pos}

        \subsection{Taggers and categories}

        Every POS-resolved analysis in this paper, the attribution of
        Figure~\ref{fig:bleuretention}, the retention decomposition of
        Figure~\ref{fig:structdiff}, the QA coverage of Appendix~\ref{app:qayield} and
        the distributions below, uses one spaCy transformer pipeline per language:
        \texttt{de\_dep\_news\_trf} for German and \texttt{zh\_core\_web\_trf} for
        Chinese, with the tagger, morphologiser and lemmatiser retained and the parser
        and NER disabled. Tokens tagged \textsc{punct} or \textsc{space} are dropped, and
        every token is reduced to its lowercased lemma, so inflectional variants of one
        word count once. Tagging is memoised on sentence text, which makes identical
        predictions across corruption conditions tagged once and identically by
        construction.

        The two-way split used throughout is content $=$ \textsc{noun}, \textsc{propn},
        \textsc{verb}, \textsc{adj}, \textsc{num}, \textsc{adv}, \textsc{pron}, and
        function $=$ \textsc{adp}, \textsc{det}, \textsc{aux}, \textsc{cconj},
        \textsc{sconj}, \textsc{part}, with any residual tag (\textsc{x}, \textsc{intj},
        \textsc{sym}) counted as function in the two-way summaries. \textsc{pron} is
        deliberately counted as content: a pronoun in SLT is a pointing sign the signer
        must actually have articulated in space, so unlike the
        \textsc{adp}/\textsc{det}/\textsc{aux} scaffolding a target-side prior cannot
        supply the correct referent for free. The $37\%$ function share of
        Section~\ref{sec:bleulimits} is not an artefact of that choice: regrouping
        \textsc{pron} as function raises it to $42.2\%$ on Phoenix-2014T and $49.6\%$ on
        CSL-Daily, so the conclusion holds, more strongly, under either convention.

        \subsection{Distribution in the test references}
        \label{app:langdist}

        \begin{figure*}[!h]
            \centering
            \begin{subfigure}[b]{0.48\linewidth}
                \centering
                \includegraphics[width=\linewidth]{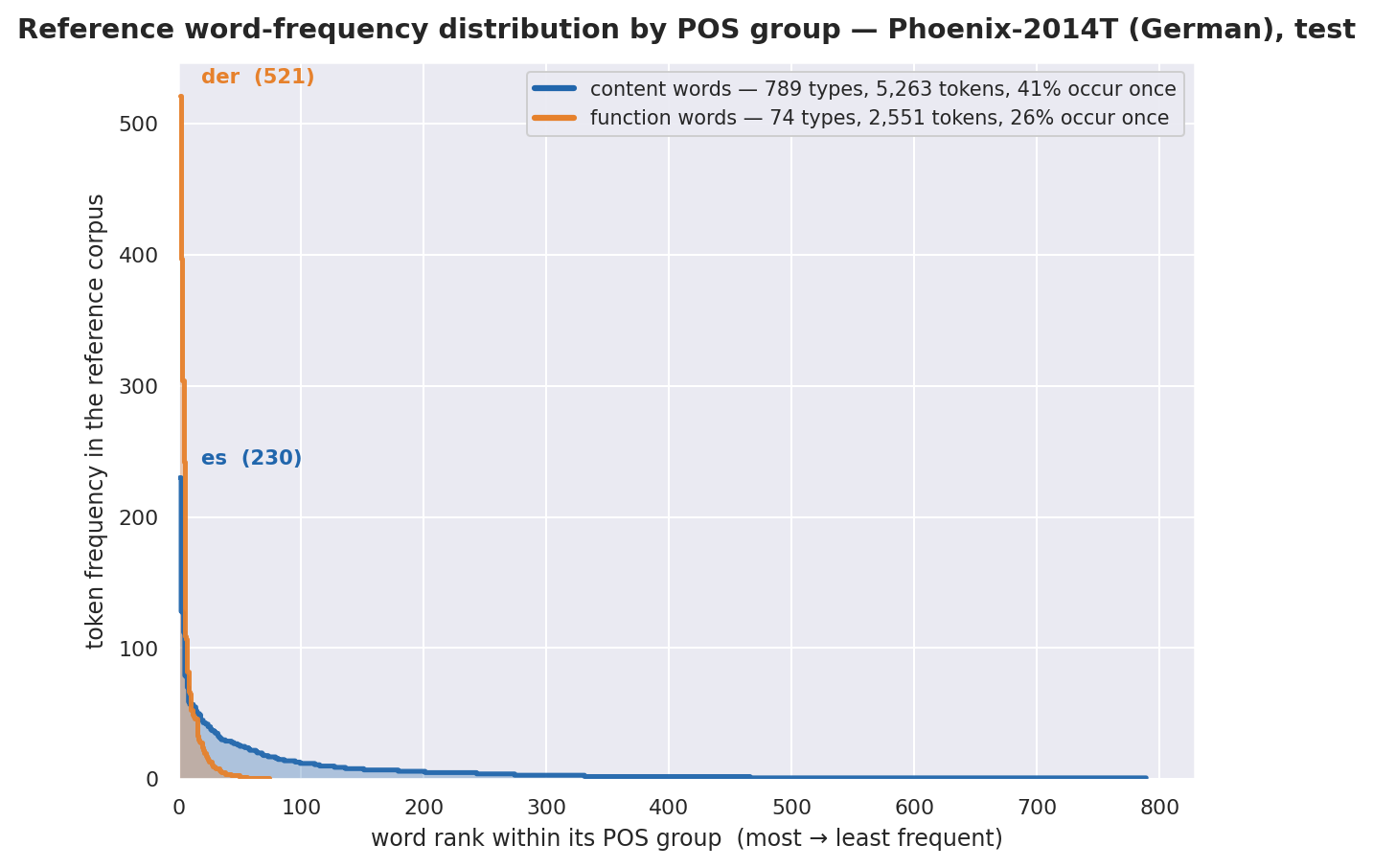}
                \caption{Phoenix-2014T, linear axes.}
                \label{fig:wordfreq_phoenix_lin}
            \end{subfigure}
            \hfill
            \begin{subfigure}[b]{0.48\linewidth}
                \centering
                \includegraphics[width=\linewidth]{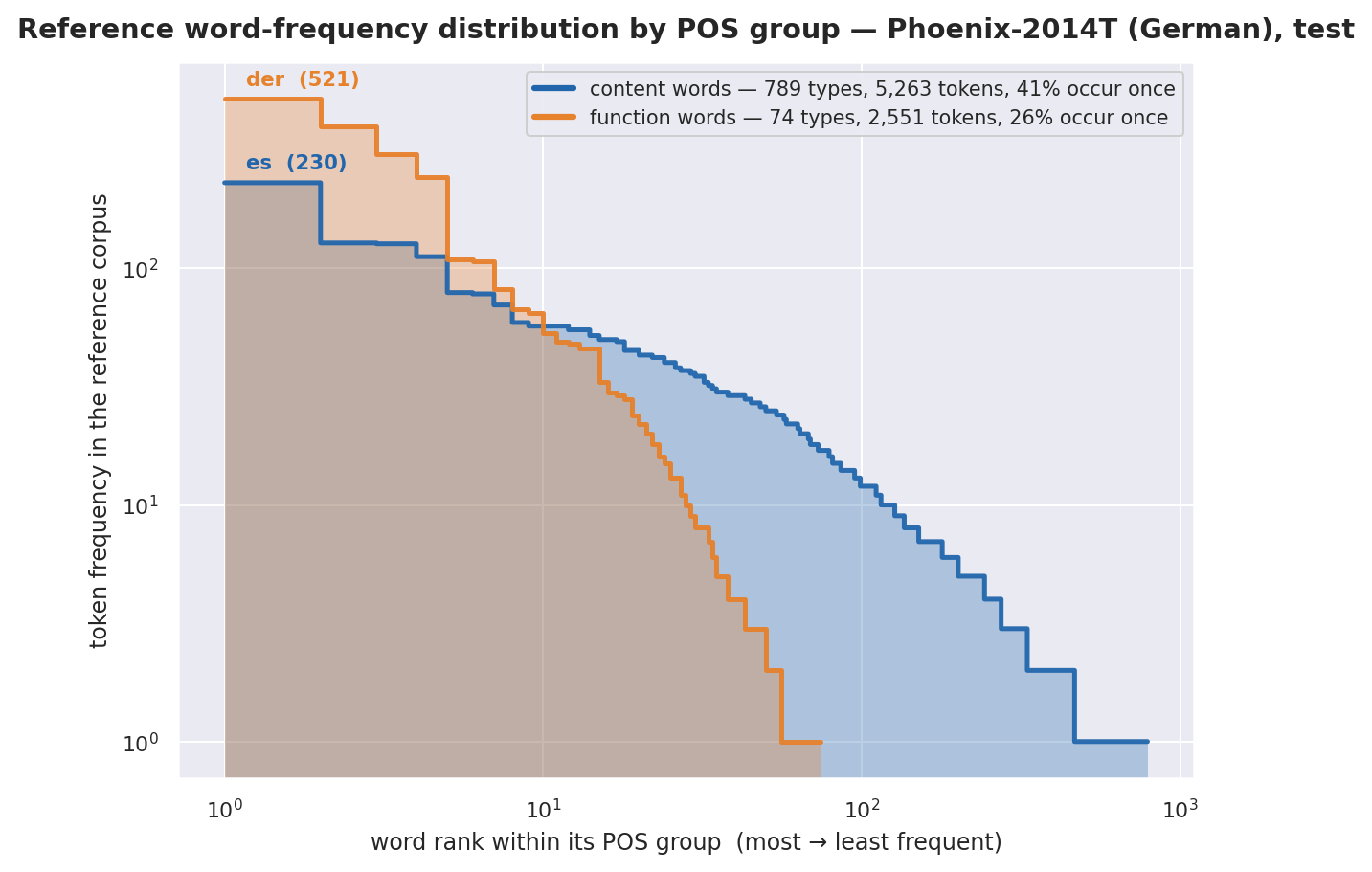}
                \caption{Phoenix-2014T, log--log axes.}
                \label{fig:wordfreq_phoenix_log}
            \end{subfigure}

            \vspace{0.5em}

            \begin{subfigure}[b]{0.48\linewidth}
                \centering
                \includegraphics[width=\linewidth]{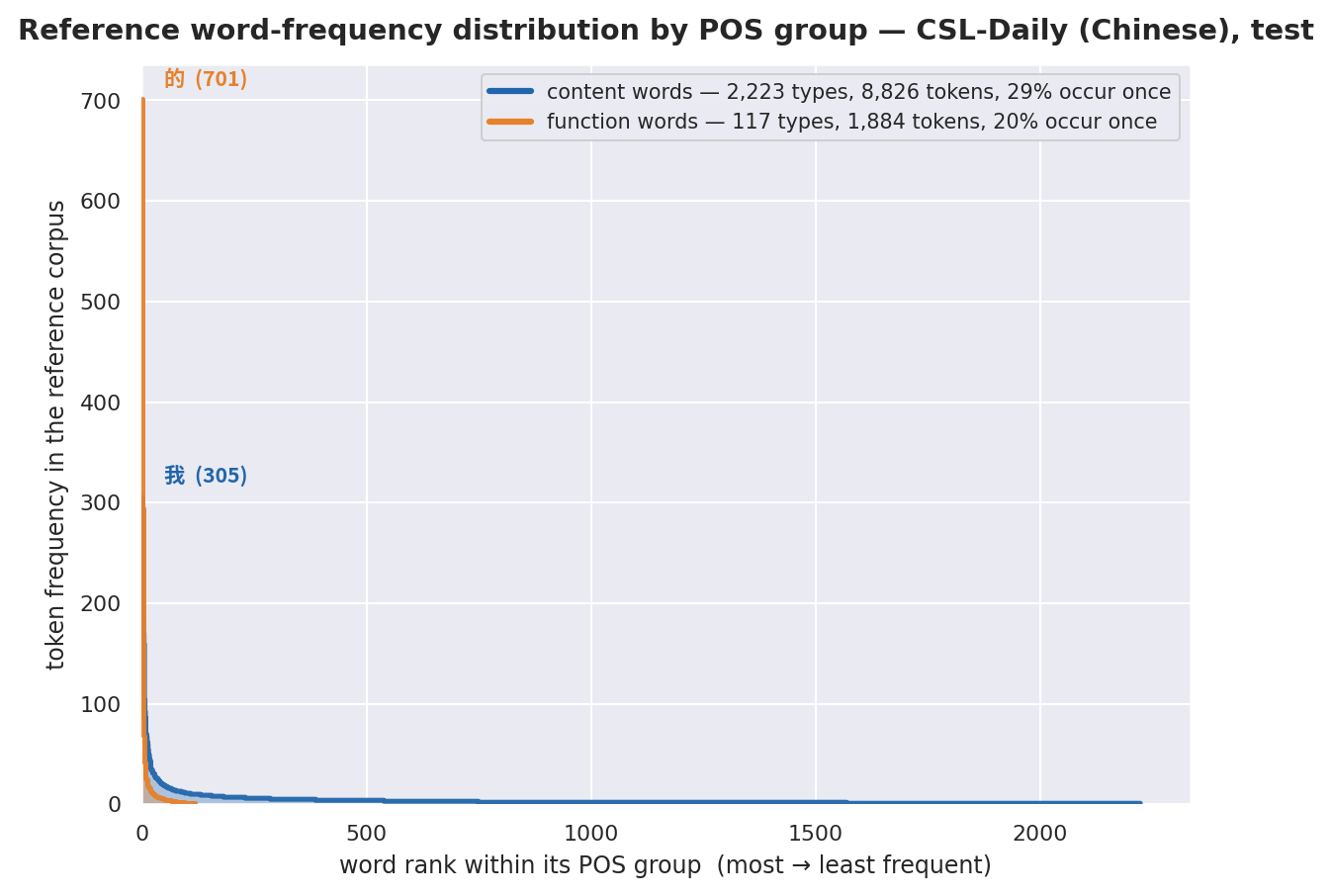}
                \caption{CSL-Daily, linear axes.}
                \label{fig:wordfreq_csl_lin}
            \end{subfigure}
            \hfill
            \begin{subfigure}[b]{0.48\linewidth}
                \centering
                \includegraphics[width=\linewidth]{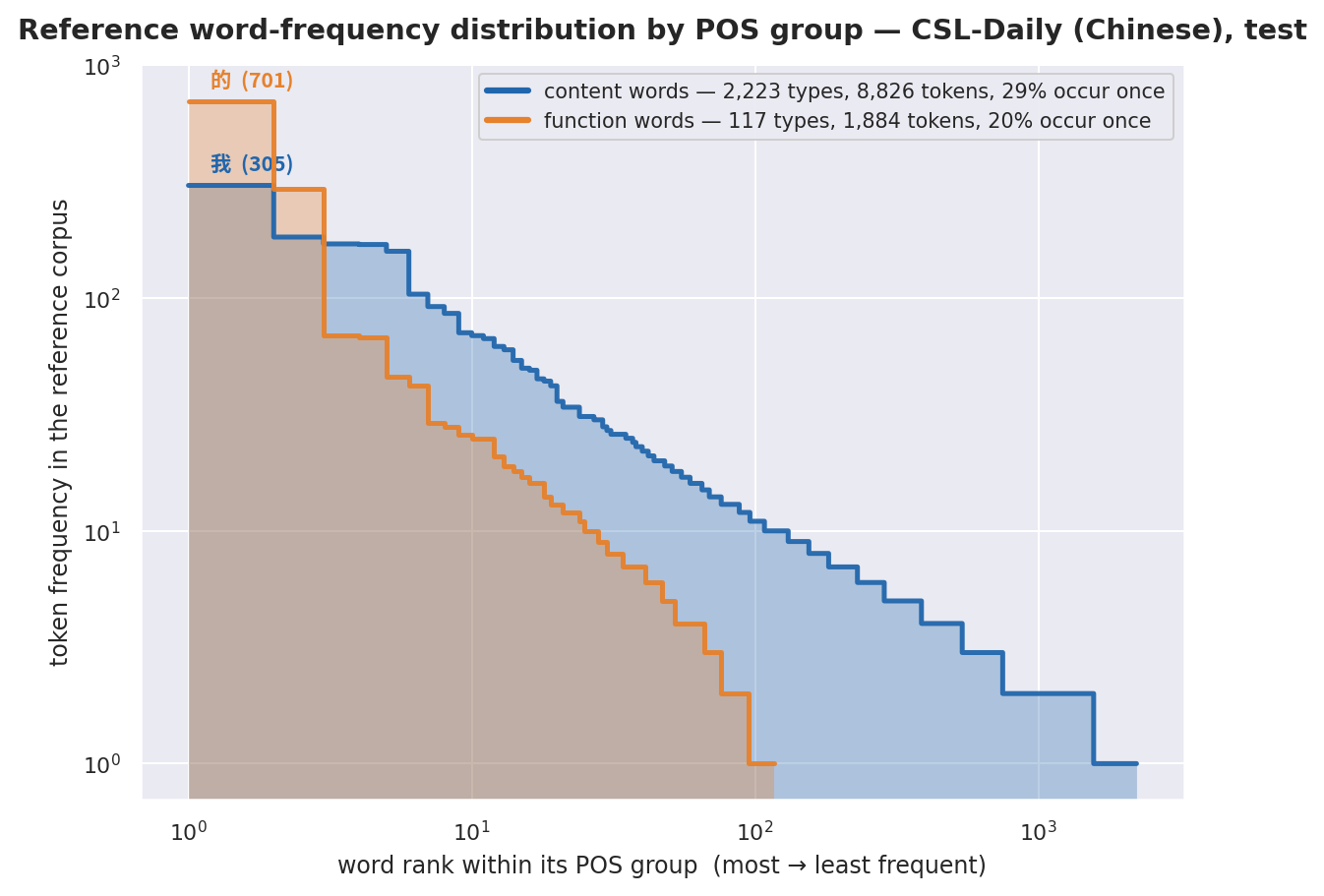}
                \caption{CSL-Daily, log--log axes.}
                \label{fig:wordfreq_csl_log}
            \end{subfigure}

            \caption{Rank-frequency curves of the spoken-language test references, word
            types sorted by descending frequency and split by the content/function
            categories above. Function words form a short, tall curve (few types, each
            frequent); content words a long, flat one with a hapax tail. Linear axes
            (left) show the concentration of function tokens, log--log axes (right) the
            length of the content tail.}
            \label{fig:wordfreq}
        \end{figure*}

        Section~\ref{sec:bleulimits} argues that function words are the words a
        target-side prior can supply for free, because they occupy a small and
        concentrated vocabulary where content words are broad and long-tailed. The test
        references bear that out directly. On Phoenix-2014T, $74$ function word types
        cover $2{,}551$ tokens ($34.5$ tokens per type) against $789$ content types over
        $5{,}263$ tokens ($6.7$ per type); the ten most frequent function types alone
        account for $76.3\%$ of all function tokens, where the ten most frequent content
        types account for $18.9\%$ of content tokens, and $41.1\%$ of content types occur
        exactly once against $25.7\%$ of function types. CSL-Daily carries a far larger
        content vocabulary relative to its function vocabulary, $19.0$ content types per
        function type against $10.7$ on Phoenix-2014T: $117$ function types over
        $1{,}884$ tokens ($16.1$ per type) against $2{,}223$ content types over $8{,}826$
        tokens ($4.0$ per type), top-ten shares of $70.5\%$ against $16.0\%$, and hapax
        rates of $19.7\%$ for function types against $29.5\%$ for content types.
        Figure~\ref{fig:wordfreq} plots the corresponding rank-frequency curves.

        A metric counting n-gram matches can therefore secure a large share of its tokens
        from a vocabulary of well under a hundred types, learnable from the target side
        alone, while the content vocabulary that the visual signal has to carry is an
        order of magnitude larger and mostly rare.

        \subsection{Transfer to gloss}

        The gloss side of Figure~\ref{fig:bleuretention} requires POS tags for annotation
        that is not running text: Phoenix-2014T glosses are uppercase German citation
        forms and CSL-Daily glosses are space-separated Chinese lemmas, so neither
        pipeline applies to them unmodified. The cached distributions cover $4{,}144$
        tagged gloss tokens on Phoenix-2014T, against $4{,}264$ whitespace-delimited
        tokens in the test split, and $4{,}551$ on CSL-Daily against $9{,}002$.


        \subsection{BLEU-4 attribution}

        The attribution distributes a system's corpus BLEU-4 over POS categories in three
        steps. First, each $n$-gram of the system output that is matched against the
        reference under BLEU's clipped counting, $n = 1 \ldots 4$, contributes $1/n$ to
        each of its $n$ tokens; a token's attribution weight is the sum of those
        contributions over every matched $n$-gram it participates in, so a token in
        several matched orders accumulates more than one and an unmatched token
        accumulates zero. Second, weights are pooled by the POS tag of the token, giving
        a per-category total $A_c$; because each matched $n$-gram distributes exactly one
        unit in total, $\sum_c A_c$ is the number of matched $n$-grams the system
        produced. Third, each category receives $A_c / \sum_{c'} A_{c'}$ of the system's
        corpus BLEU-4. The assignment is exact by construction: the per-category points
        sum to the corpus BLEU-4 of the system they were computed for, to four decimals,
        for all six systems on both corpora. The cached table additionally records a
        match rate per category, the fraction of that category's tokens participating in
        at least one matched $n$-gram, which separates how often a category is hit from
        how much credit it carries.
        
        \section{QA Generation: Implementation Details}
        \label{app:prompts}
    
        \paragraph{Teacher configuration.} One open-weight LLM plays every role (content
        extraction, question generation, distractor repair, all quality-control gates,
        and answering), served locally through an OpenAI-compatible vLLM endpoint.
        Decoding is greedy ($\tau=0$) except for the two recall paths: the fallback retry
        of content extraction ($\tau=0.7$) and the second per-content question-generation
        pass ($\tau=0.7$). Generation calls are given an 8{,}192-token budget to
        accommodate long reasoning traces before the JSON payload; answering calls are
        given 1{,}024. Reasoning effort is set to \texttt{low} for reasoning-capable
        checkpoints, which avoids reasoning overflowing the budget and returning empty
        content. All model outputs are parsed by extracting the first balanced JSON
        object, and every call is memoised in an append-only cache keyed on (backend,
        call type, model, prompt hash, input, token budget, reasoning effort,
        temperature), making runs resumable and bit-identical on repetition.
    
        \paragraph{Pipeline stages.} For a reference $s$:
        \begin{enumerate}[nosep,leftmargin=*]
            \item \textbf{Content unit extraction.} $s\rightarrow$ typed checkable content units; retry at $\tau=0.7$ if empty; deduplicate on normalised text. An optional audit pass that re-checks each content unit for groundedness is implemented but disabled in the reported runs.
            \item \textbf{Question generation.} Each content unit independently $\rightarrow$ 2--4 MCQs, over two temperature passes, unioned under deduplication on both normalised question and normalised answer.
            \item \textbf{Distractor repair.} Items without exactly four distractors get one repair call; still-malformed items are dropped.
            \item \textbf{Structural validation.} Exactly four non-empty, mutually distinct distractors, none equal to the answer; the answer must not appear inside the question; truncation to $N_Q^{\max}=10$.
            \item \textbf{Assembly.} Five content options permuted under a seed derived from \textsc{sha256}(reference id, question, answer); the language's \emph{not stated} sentinel appended as option~5.
            \item \textbf{Gates.} Round-trip, empty-passage world-knowledge probe, and single-answer ambiguity check; a fourth gibberish gate is available under strict mode and unused here.
            \item \textbf{Answering.} Given only the passage, question and six options, the answerer is instructed to rely solely on the passage, accept paraphrase as a match, select \emph{not stated} when the passage is silent, and emit a single digit, read as the last standalone in-range digit of the completion.
        \end{enumerate}
        If a reference survives extraction but yields no validated question, the pipeline
        falls back to a single folded generation prompt before the reference is dropped.
    
        \paragraph{Language resolution.} Prompts are resolved in the domain language of
        the dataset or partition: the extractor, question generator, answerer, fallback
        generator, \emph{not stated} sentinel, and all questions, answers and distractors
        are written in the reference language. The cross-lingual comparisons
        (OpusParcus, PAWS-X) instead use one language-agnostic prompt set across all
        languages, so that no language benefits from prompt tuning. BLEU is computed with
        a \textsc{sacreBLEU} tokenizer throughout.
    
        \paragraph{Prompts.} Verbatim listings of the extractor, content-to-question
        generator, answerer, folded fallback generator, and the three auxiliary gate
        prompts (ambiguity, distractor repair, content unit audit) are provided in the
        supplementary material. The \emph{not stated} sentinel is registered per language
        (e.g.\ \textit{nicht genannt} for German, \textit{ei mainittu} for Finnish,
        \textit{non mentionn\'e} for French), currently covering the ten languages used
        in this work.

         \section{Metric Definitions}
        \label{app:metricdefs}
    
                \subsection{Per-item metric scores}
        For a reference string $r$ scored against a source string $s$, both metrics are
        mapped to $[0,1]$:
        $$
        b(r,s) = \frac{\text{sacreBLEU-4}\bigl(r,[s]\bigr)}{100},
        $$
        $$
        q(r,s) = \frac{1}{|Q_s|}\sum_{Q\in Q_s}
                \mathbf{1}\bigl[\text{ans}(Q,r)=\text{gold}(Q)\bigr]
        $$
        where $Q_s$ is the set of QC-gated multiple-choice questions generated from the
        source rendering $s$ (each with a gold option), and $\text{ans}(Q,r)$ is the
        teacher's chosen option when shown reference $r$.

        BLEU-4 is \textsc{sacreBLEU} throughout, under the tokeniser matched to the target
        language: \texttt{zh} for Chinese, character-level for Japanese and Korean, and
        \texttt{13a} otherwise, since the default \texttt{13a} tokeniser mangles CJK and
        would put those languages on a different footing from the rest. No lowercasing or
        punctuation stripping is applied anywhere.

        \subsection{Confidence intervals}
        \label{app:ci}
        Every QA interval in this paper is a percentile bootstrap over test instances.
        Let $a_i\in[0,1]$ be the fraction of instance $i$'s admitted questions that a
        system answers correctly, so the reported score is $\bar a = N^{-1}\sum_{i=1}^{N}
        a_i$ over the $N$ references retaining at least one question. We draw $B=2{,}000$
        resamples of the $N$ instances with replacement, recompute $\bar a$ on each, and
        take the $2.5$th and $97.5$th percentiles; a half-width is
        $(\text{hi}-\text{lo})/2$. The draw is seeded, so an interval is reproducible
        from the stored per-instance accuracies.
    
        The resampling unit is the instance, not the question. A reference contributes
        several questions ($5.2$--$6.4$ on average,
        Table~\ref{tab:questiongeneration}) whose correctness is correlated through the
        shared reference and the shared candidate sentence, so resampling questions
        independently would treat that correlation as extra evidence and report an
        interval narrower than the data supports.
    
        This interval covers the finite test set only: it asks how far the score would
        move on another sample of sentences with the bank held fixed. It does not cover
        regeneration of the bank, which Section~\ref{sec:cost} reports separately as
        $\sigma$ over ten banks. The two are not directly comparable as printed, since
        $\sigma$ is one standard deviation while CI/2 is a $95\%$ half-width, i.e.\ about
        $1.96$ standard errors; on a common standard-error basis the instance term is
        three to nine times the bank term, depending on the system.

        The BLEU-4 intervals of Table~\ref{tab:combined_results} are produced the same
        way, with the resampled quantity changed: each of the $B=2{,}000$ resamples draws
        the same $N$ instances with replacement and recomputes corpus BLEU-4 over the
        resampled hypothesis and reference pairs under the language's \textsc{sacreBLEU}
        tokeniser, and the interval is
        again the $2.5$th to $97.5$th percentile with half-width
        $(\mathrm{hi}-\mathrm{lo})/2$. Both columns of that table therefore share one
        resampling unit and one resample count, and neither covers regeneration of the
        question bank. The asymmetry is that this is the only source of variation for
        BLEU-4, which is deterministic given a checkpoint, whereas the QA score carries
        the bank term of Section~\ref{sec:cost} in addition.

        \subsection{Signal-to-noise ratio (SNR)}
        Write $c(\cdot)$ for either metric of Section~\ref{app:metricdefs}, $b$ or $q$.
        Data are organized into groups $g=1,\dots,G$, each a set of equivalent renderings
        of one source sentence with a designated source rendering $s_g$.
    
        \paragraph{Paraphrase scores.} The other renderings of the same sentence, scored
        against $s_g$ (the metric's behaviour on genuine paraphrases):
        $$
        W_g = \{\, c(r, s_g) : r \in \text{group } g,\ r \neq s_g \,\},
        \qquad
        n_g = |W_g|.
        $$
        Per-group mean and population standard deviation (SD):
        $$
        \mu_g = \frac{1}{n_g}\sum_{x\in W_g} x,
        $$
        $$
        \sigma_g = \sqrt{\frac{1}{n_g}\sum_{x\in W_g}(x-\mu_g)^2}
        \quad(\sigma_g=0 \text{ if } n_g=1).
        $$
    
        \paragraph{Floor scores.} Renderings of \emph{different} sentences scored against
        $s_g$, pooled over all groups (the metric's floor on non-equivalent content):
        $$
        F = \{\, c(r', s_g) : r' \notin \text{group } g \,\}.
        $$
    
        \paragraph{Aggregate over groups}
        \[
        \underbrace{\bar\mu = \tfrac{1}{G}\textstyle\sum_{g}\mu_g}_{\text{paraphrase mean}},
        \qquad
        \underbrace{\bar\sigma = \tfrac{1}{G}\textstyle\sum_{g}\sigma_g}_{\text{paraphrase SD}},
        \]
        \[
        \underbrace{\bar f = \tfrac{1}{|F|}\textstyle\sum_{x\in F} x}_{\text{floor mean}}.
        \]
        \[
        \boxed{\;
        \text{gap}\ \Delta = \bar\mu - \bar f,
        \qquad
        \mathrm{SNR} = \frac{\Delta}{\bar\sigma},
        \qquad
        \mathrm{NSR} = \frac{1}{\mathrm{SNR}}
        \;}
        \]
    
        \paragraph{Interpretation.} $\bar\sigma$ is the average dispersion within a set of
        equivalent phrasings (paraphrase fluctuation, the \emph{noise}); $\Delta$ is how
        far the metric moves between equivalent and non-equivalent content (the
        discrimination \emph{signal}). SNR is dimensionless (numerator and denominator
        share the $[0,1]$ scale), so it is comparable across BLEU and QA, unlike raw
        $\bar\sigma$, which a floor-bound metric makes trivially small.
        \textbf{Higher SNR = more paraphrase-invariant} (equivalently, lower NSR). Because
        both metrics floor on non-equivalent content ($\bar f \approx 0$), $\Delta \approx
        \bar\mu$ and SNR coincides with the coefficient-of-variation signal-to-noise
        $\bar\mu/\bar\sigma$, while remaining well defined for metrics that do not floor.
        (Defined when $\Delta>0$.)
    
        \paragraph{Auxiliary statistics.} Discriminability
        $d = \Delta / \operatorname{sd}\bigl(\{\mu_g\}\cup F\bigr)$; significance via a
        paired one-sided Wilcoxon signed-rank test on
        $\{(\sigma_g^{\mathrm{BLEU}},\sigma_g^{\mathrm{QA}})\}$ over groups with
        $n_g\ge2$.

                \section{Human Evaluation of QA-based pipeline}
        \label{app:humaneval}
     To assess the reliability of the automated QA-generation pipeline, human evaluations are conducted for the content span extraction, question–answer quality assessment, and distractor validity. A annotator rate a samples, and report the acceptance rate alongside its Wilson 95\% confidence interval. Table~\ref{tab:worked_example} presents representative question–answer sets produced at the final stage of the pipeline.

    \begin{table*}[!t]
        \centering
        \small
        \setlength{\tabcolsep}{4pt}
        \caption{Sample references carried through the pipeline. Within each
        item the correct answer is marked $\star$ and set in bold; four of the remaining
        options are distractors and the last is a \emph{not stated} sentinel.}
        \label{tab:worked_example}
        \begin{tabular}{p{0.25\linewidth} p{0.67\linewidth}}
            \toprule
            \textbf{Content unit} & \textbf{Question and answer options} \\
            \midrule
            \multicolumn{2}{p{0.94\linewidth}}{\textbf{Phoenix-2014T (de)} ---
              es kommen dann ein paar wolkenfelder \newline
              \emph{Then a few cloud fields come.}} \\
            \midrule
            kommen \emph{(come)} &
              Was geschieht mit den Wolkenfeldern? \emph{(what happens with the cloud fields?)} \newline
              ziehen fort \emph{move away} $\cdot$ bleiben stehen \emph{stay standing}
              $\cdot$ verdampfen \emph{evaporate} $\cdot$ teilen sich \emph{divide themselves}
              $\cdot$ $\star$\,\textbf{kommen \emph{come}}
              $\cdot$ nicht genannt \emph{not mentioned} \\
            \addlinespace
            ein paar \emph{(a few)} &
              Wie viele Wolkenfelder kommen? \emph{(how many cloud fields come?)} \newline
              viele \emph{many} $\cdot$ mehrere \emph{several}
              $\cdot$ $\star$\,\textbf{ein paar \emph{a few}} $\cdot$ eins \emph{one}
              $\cdot$ keine \emph{no} $\cdot$ nicht genannt \emph{not mentioned} \\
            \addlinespace
            wolkenfelder \emph{(cloud fields)} &
              Was kommt nachher? \emph{(what comes later?)} \newline
              sonnenschein \emph{sunshine} $\cdot$ sturm \emph{storm} $\cdot$ regen \emph{rain}
              $\cdot$ $\star$\,\textbf{wolkenfelder \emph{cloud fields}}
              $\cdot$ niederschlag \emph{precipitation}
              $\cdot$ nicht genannt \emph{not mentioned} \\
            \midrule
            \multicolumn{2}{p{0.94\linewidth}}{\textbf{CSL-Daily (zh)} ---
              \zh{他的膝关节手术非常成功。} \newline
              \emph{His knee surgery was very successful.}} \\
            \midrule
            \raggedright \zh{他} \emph{(he)} &
              \zh{谁的膝关节手术非常成功？} \emph{(whose knee surgery was very successful?)} \newline
              \zh{护士} \emph{nurse} $\cdot$ $\star$\,\textbf{\zh{他} \emph{he}} $\cdot$ \zh{她} \emph{she}
              $\cdot$ \zh{医生} \emph{doctor} $\cdot$ \zh{病人} \emph{patient} $\cdot$ \zh{未提及} \emph{not mentioned} \\
            \addlinespace
            \raggedright \zh{膝关节手术} \emph{(knee surgery)} &
              \zh{什么手术非常成功？} \emph{(what surgery was very successful?)} \newline
              \zh{心脏手术} \emph{heart surgery} $\cdot$ \zh{眼科手术} \emph{eye surgery}
              $\cdot$ \zh{脑部手术} \emph{brain surgery}
              $\cdot$ $\star$\,\textbf{\zh{膝关节手术} \emph{knee surgery}}
              $\cdot$ \zh{骨盆手术} \emph{pelvic surgery} $\cdot$ \zh{未提及} \emph{not mentioned} \\
            \addlinespace
            \raggedright \zh{非常成功} \emph{(very successful)} &
              \zh{手术的结果如何？} \emph{(how was the result of the surgery?)} \newline
              \zh{还需要再做一次} \emph{needs doing again} $\cdot$ \zh{失败了} \emph{it failed}
              $\cdot$ \zh{有并发症} \emph{there were complications}
              $\cdot$ $\star$\,\textbf{\zh{非常成功} \emph{very successful}}
              $\cdot$ \zh{一般般} \emph{so-so} $\cdot$ \zh{未提及} \emph{not mentioned} \\
            \midrule
            \multicolumn{2}{p{0.94\linewidth}}{\textbf{PAWS-X (en)} --- The Tabaci River is a
              tributary of the River Leurda in Romania~.} \\
            \midrule
            Tabaci River & What river is a tributary of the River Leurda? \newline
              $\star$\,\textbf{Tabaci River} $\cdot$ Leurda River $\cdot$ Danube River $\cdot$ Olt River
              $\cdot$ Mures River $\cdot$ not stated \\
            \addlinespace
            Tabaci River & In which country is the Tabaci River located? \newline
              $\star$\,\textbf{Romania} $\cdot$ Hungary $\cdot$ Ukraine $\cdot$ Serbia $\cdot$ Bulgaria $\cdot$ not stated \\
            \addlinespace
            tributary & What is the Tabaci River? \newline
              $\star$\,\textbf{tributary} $\cdot$ main river $\cdot$ lake $\cdot$ canal $\cdot$ waterfall $\cdot$ not stated \\
            \addlinespace
            tributary & Of which river is the Tabaci River a tributary? \newline
              $\star$\,\textbf{River Leurda} $\cdot$ River Someş $\cdot$ River Danube $\cdot$ River Amara
              $\cdot$ River Olt $\cdot$ not stated \\
            \addlinespace
            tributary & What is the relationship of the Tabaci River to the River Leurda? \newline
              $\star$\,\textbf{tributary} $\cdot$ source $\cdot$ mainstream $\cdot$ confluence $\cdot$ branch $\cdot$ not stated \\
            \addlinespace
            River Leurda & Which river is the Tabaci River a tributary of? \newline
              $\star$\,\textbf{River Leurda} $\cdot$ River Someş $\cdot$ Danube River $\cdot$ River Timiş
              $\cdot$ River Olt $\cdot$ not stated \\
            \addlinespace
            River Leurda & In which country is the River Leurda located? \newline
              $\star$\,\textbf{Romania} $\cdot$ Ukraine $\cdot$ Serbia $\cdot$ Hungary $\cdot$ Bulgaria $\cdot$ not stated \\
            \addlinespace
            Romania & In which country is the Tabaci River a tributary of the River Leurda? \newline
              $\star$\,\textbf{Romania} $\cdot$ Serbia $\cdot$ Bulgaria $\cdot$ Ukraine $\cdot$ Hungary $\cdot$ not stated \\
            \bottomrule
        \end{tabular}
    \end{table*}
    
        \subsection{Evaluation of content unit extraction}
        
        A \emph{content unit} is a span copied verbatim from the reference that names an isolated component of the expression targeting an entity or group, an action, an attribute, a quantity, a time, a location, a relation, a negation, or a polarity. Prepositions and modifiers may stay inside the phrase they belong to, so \emph{on the mountains}, \emph{it lightly rains} and \emph{from east to northeast} may be considered one unit. Articles, copulas and conjunctions should not be units on their own. Given the reference and the extracted units, the judge marks: \textbf{grounding}: Is each extracted content unit present in the reference; \textbf{missing}: Is any content from the reference not covered by any extracted unit; and \textbf{excess}: Could any of the extracted units be omitted without losing content.
        
        The evaluation covers 50 spans from 4 datasets (Phoenix-2014T, csl-daily, Opus-Parcus and WMT23), covering 3 languages. Chinese and German is translated to english before it is evaluated. The results are presented in Table \ref{tab:human_eval}. Generally, the LLM appears perfectly able to consistently produce grounded spans. When assessing the span size, it appears appropriate in roughly 80\% of the cases, with some tendencies to over- and under-segment spans. Undersegmentation appears only in the longest references, which are the most frequent in the WMT news sentences. Over segmentation are concentrated in the short references, more frequently occuring in Phoenix-2014T. While some structural words are ocationally passed as content units, their impact on the question generation phase appears limited. A low generall rate for missed content is observed.

    \begin{table*}[!t]
    \centering
    \scriptsize
    \setlength{\tabcolsep}{4pt}
    \caption{Human evaluation of the question-generation pipeline by one judge. Each dataset is capped at its first 100 judged spans and 50 judged questions and items, in annotation order, so every corpus carries equal weight in the pooled column. Subscripts are Wilson 95\% intervals: for a rate that should be high ($\uparrow$) the lower bound is the conservative reading, for one that should be low ($\downarrow$) the upper bound is. Each block states its own denominator, since spans, content words, questions and items are four different bases. \emph{Usable} deliberately excludes whether a question is anchored on its own unit, which is a property of the extracted span rather than of the question; that code is reported separately.}
    \label{tab:human_eval}
    \begin{tabular}{l|c|cccc}
    \toprule
    Check & \textbf{All} & Phoenix (de) & CSL (zh) & OPUS-P (en) & WMT23 (zh-en) \\
    \midrule
    \multicolumn{6}{l}{\textbf{Content-unit extraction}} \\
    \cmidrule(l){1-6}
      spans judged & \textbf{400} & 100 & 100 & 100 & 100 \\
      Grounding $\uparrow$ & \textbf{$100.0_{[99.0,\,100.0]}$} & $100.0_{[96.3,\,100.0]}$ & $100.0_{[96.3,\,100.0]}$ & $100.0_{[96.3,\,100.0]}$ & $100.0_{[96.3,\,100.0]}$ \\
      Excess $\downarrow$ & \textbf{$5.0_{[3.3,\,7.6]}$} & $8.0_{[4.1,\,15.0]}$ & $6.0_{[2.8,\,12.5]}$ & $5.0_{[2.2,\,11.2]}$ & $1.0_{[0.2,\,5.4]}$ \\
      reference content words & \textbf{793} & 139 & 244 & 134 & 276 \\
      Missed $\downarrow$ & \textbf{$0.9_{[0.4,\,1.8]}$} & $0.0_{[0.0,\,2.7]}$ & $0.4_{[0.1,\,2.3]}$ & $3.0_{[1.2,\,7.4]}$ & $0.7_{[0.2,\,2.6]}$ \\
    \midrule
    \multicolumn{6}{l}{\textbf{Question generation}} \\
    \cmidrule(l){1-6}
      questions judged & \textbf{200} & 50 & 50 & 50 & 50 \\
      Usable $\uparrow$ & \textbf{$96.0_{[92.3,\,98.0]}$} & $98.0_{[89.5,\,99.6]}$ & $94.0_{[83.8,\,97.9]}$ & $96.0_{[86.5,\,98.9]}$ & $96.0_{[86.5,\,98.9]}$ \\
      Answerable $\uparrow$ & \textbf{$98.0_{[95.0,\,99.2]}$} & $100.0_{[92.9,\,100.0]}$ & $94.0_{[83.8,\,97.9]}$ & $100.0_{[92.9,\,100.0]}$ & $98.0_{[89.5,\,99.6]}$ \\
      Grounded binding $\uparrow$ & \textbf{$98.5_{[95.7,\,99.5]}$} & $100.0_{[92.9,\,100.0]}$ & $94.0_{[83.8,\,97.9]}$ & $100.0_{[92.9,\,100.0]}$ & $100.0_{[92.9,\,100.0]}$ \\
      Well formed $\uparrow$ & \textbf{$96.5_{[93.0,\,98.3]}$} & $98.0_{[89.5,\,99.6]}$ & $94.0_{[83.8,\,97.9]}$ & $96.0_{[86.5,\,98.9]}$ & $98.0_{[89.5,\,99.6]}$ \\
      Anchored on its own unit $\uparrow$ & \textbf{$98.0_{[95.0,\,99.2]}$} & $96.0_{[86.5,\,98.9]}$ & $98.0_{[89.5,\,99.6]}$ & $98.0_{[89.5,\,99.6]}$ & $100.0_{[92.9,\,100.0]}$ \\
    \midrule
    \multicolumn{6}{l}{\textbf{Answer options}} \\
    \cmidrule(l){1-6}
      items judged & \textbf{200} & 50 & 50 & 50 & 50 \\
      Guessable closed book $\downarrow$ & \textbf{$15.0_{[10.7,\,20.6]}$} & $16.0_{[8.3,\,28.5]}$ & $14.0_{[7.0,\,26.2]}$ & $14.0_{[7.0,\,26.2]}$ & $16.0_{[8.3,\,28.5]}$ \\
      Item validity $\uparrow$ & \textbf{$98.0_{[95.0,\,99.2]}$} & $94.0_{[83.8,\,97.9]}$ & $100.0_{[92.9,\,100.0]}$ & $100.0_{[92.9,\,100.0]}$ & $98.0_{[89.5,\,99.6]}$ \\
      Key correct $\uparrow$ & \textbf{$99.5_{[97.2,\,99.9]}$} & $100.0_{[92.9,\,100.0]}$ & $100.0_{[92.9,\,100.0]}$ & $98.0_{[89.5,\,99.6]}$ & $100.0_{[92.9,\,100.0]}$ \\
      Sound distractors $\uparrow$ & \textbf{$95.8_{[94.1,\,96.9]}$} & $95.0_{[91.0,\,97.3]}$ & $97.5_{[94.3,\,98.9]}$ & $98.5_{[95.7,\,99.5]}$ & $92.0_{[87.4,\,95.0]}$ \\
      Clean item $\uparrow$ & \textbf{$90.5_{[85.6,\,93.8]}$} & $88.0_{[76.2,\,94.4]}$ & $94.0_{[83.8,\,97.9]}$ & $94.0_{[83.8,\,97.9]}$ & $86.0_{[73.8,\,93.0]}$ \\
    \bottomrule
    \end{tabular}
    \end{table*}

        \subsection{Question-answer generation}
        The judge is shown the reference, the source unit and the generated question. Four properties are marked: whether the question tests that unit, whether it is answerable from the reference alone, whether it assumes only relations the reference actually states, and whether it is well formed. A question counts as usable only if all four hold.

        \subsection{Distractor generation}
        The key is disclosed and each of the four distractors is coded as clearly wrong but plausible, also correct, a synonym of the key, the wrong kind of thing so that it does not answer the question at all, or the right kind but too implausible to compete. The wrong-kind code is kept separate because four options of the wrong kind leave the key as the only one that fits, which is what makes an item solvable without reading anything.
        Two passes, in this order. Closed book, the judge sees the question and the six options with no reference shown at all and picks the option they would bet on; chance is $1/6$. Since every admitted item is one the model itself could not answer with the passage withheld, an item a judge can answer is a residual miss of that check. The reference is then revealed with the key still hidden and the judge answers the item; agreement with the key is item validity measured without anchoring.
    
        \subsection{Answering step}
        
    \begin{table*}[!t]
\centering
\small
\setlength{\tabcolsep}{4pt}
\caption{Human--model agreement on the answering step, with key and model
choice withheld from the judge. \emph{Reference passage} is a control: items
are answerable by construction, so it measures whether a person recovers the
same answer from text known to contain it. \emph{System translation} is the
metric's actual task, on unseen decoder output. Subscripts are Wilson 95\%
intervals; $\kappa$ is Cohen's $\kappa$ over the six options.}
\label{tab:human_answer}
\begin{tabular}{l|ccc|ccc}
\toprule
 & \multicolumn{3}{c|}{Reference passage} & \multicolumn{3}{c}{System translation} \\
\cmidrule(lr){2-4}\cmidrule(lr){5-7}
Dataset & $n$ & Agreement & $\kappa$ & $n$ & Agreement & $\kappa$ \\
\midrule
  \textbf{Average} & 50 & \textbf{$100.0_{[92.9,\,100.0]}$} & \textbf{1.00} & 50 & \textbf{$92.0_{[81.2,\,96.8]}$} & \textbf{0.88} \\
\midrule
  Phoenix-2014T (de) & 25 & $100.0_{[86.7,\,100.0]}$ & 1.00 & 25 & $88.0_{[70.0,\,95.8]}$ & 0.85 \\
  CSL-Daily (zh) & 25 & $100.0_{[86.7,\,100.0]}$ & 1.00 & 25 & $96.0_{[80.5,\,99.3]}$ & 0.89 \\
\bottomrule
\end{tabular}
\end{table*}

    To assess the answering step against human judgment, one judge answers 50
items each from Phoenix-2014T and CSL-Daily under two conditions: the
reference passage (a control, since admitted items are answerable by
construction) and real gloss-free decoder output on unseen references (the
actual task). Table~\ref{tab:human_answer} reports agreement.
    
On reference passages, judge and model agreed on all 50 items and both
matched the pipeline's key on all 50. On
system translations, agreement was $92.0\%$ ($\kappa=0.88$), with the two
datasets differing in answerability rather than agreement: the judge could
answer 13 of 25 Phoenix items from the translation but only 3 of 25 for
CSL-Daily, where the baseline preserves little reference content. That
agreement remains high even as answerability collapses supports the intent of
the control: disagreements reflect what the translations convey, not how the
items are read.

All four disagreements take the same form: the judge selected \emph{not stated},
treating the information as unavailable from the translation, where the model
committed to an option; in no case did both commit to different answers. Taking
the judge's reading as correct, all four are over-credits. In three of them the
model reproduced the pipeline's key even though the translation at best vaguely state
it:  snow \emph{down to middle elevations}, rendered as \emph{above about four hundred
metres}, was credited as \emph{up to high elevations}. The judge selected
\emph{not stated} more often than the model in both datasets ($28$ vs.\ $24$
pooled), so the residual disagreement is directional: the metric is marginally
more generous than a human reader.

        \section{Per-language Question Yield and Content Coverage}
        \label{app:qayield}
    
        Table~\ref{tab:questiongeneration} of Section~\ref{sec:qatext} pools question
        yield and content coverage over languages; Table~\ref{tab:qayield} gives the
        per-language breakdown. Coverage is computed with the taggers of
        Appendix~\ref{app:pos}, using spaCy's \emph{large} pipeline for every language
        except German, which keeps the \texttt{de\_dep\_news\_trf} transformer model of
        the POS analysis so the SLT numbers remain directly comparable with it. PAWS-X
        Korean is scored by the metric but omitted here, as no spaCy pipeline is
        available for it.
    
        The complementary measurement to item-level coverage is what the gold answers
        themselves assert, since a candidate is only ever scored through an answer. Here
        the asymmetry the metric is designed for appears directly: on Phoenix-2014T the
        answers carry $66.6\%$ of the reference's content lemmas against $51.9\%$ of its
        function lemmas, and the gap widens on the longer PAWS-X sentences ($64.3\%$
        vs.\ $32.5\%$). Per POS (Table~\ref{tab:qaposcov}), gold answers recover $93.5\%$
        of numerals, $80.8\%$ of adjectives and $76.8\%$ of nouns on Phoenix-2014T, but
        only $18.2\%$ of auxiliaries and $8.3\%$ of subordinating conjunctions. The QA
        metric therefore concentrates its scoring on precisely the categories the
        attribution study of Section~\ref{sec:bleulimits} showed BLEU-4 to under-weight
        relative to the signed input.
    
        \begin{table}[!h]
        \centering\small
        \setlength{\tabcolsep}{3pt}
   \begin{tabular}{llrrrr}
    \toprule
    & & & & \multicolumn{2}{c}{\textbf{Content cov.}} \\
    \cmidrule(lr){5-6}
    \textbf{Benchmark} & \textbf{Lg} & $N_{\text{ref}}$ & $|Q|$ & item & answer \\
    \midrule
    Phoenix-2014T & de & 641 & 5{,}934 & $88.9_{\pm 13.6}$ & $74.2_{\pm 17.6}$  \\
    CSL-Daily & zh & 1{,}175 & 7{,}992 & $94.5_{\pm 10.7}$ & $81.7_{\pm 18.3}$  \\
    \midrule
    OpusParcus & en & 2{,}772 & 7{,}022 & $80.6_{\pm 25.0}$ & $70.8_{\pm 27.2}$  \\
    OpusParcus & de & 2{,}997 & 9{,}088 & $84.8_{\pm 22.4}$ & $68.6_{\pm 27.0}$  \\
    OpusParcus & fi & 3{,}380 & 11{,}245 & $83.9_{\pm 24.7}$ & $74.1_{\pm 27.9}$ \\
    OpusParcus & fr & 3{,}069 & 9{,}822 & $81.2_{\pm 23.9}$ & $64.4_{\pm 27.0}$  \\
    OpusParcus & ru & 3{,}631 & 12{,}450 & $90.0_{\pm 20.6}$ & $77.0_{\pm 25.1}$  \\
    OpusParcus & sv & 3{,}643 & 11{,}805 & $87.1_{\pm 22.2}$ & $70.0_{\pm 27.0}$  \\
    \midrule
    PAWS-X & en & 1{,}995 & 19{,}547 & $95.8_{\pm 8.5}$ & $83.4_{\pm 15.6}$  \\
    PAWS-X & de & 1{,}969 & 22{,}013 & $95.2_{\pm 8.8}$ & $78.8_{\pm 16.3}$  \\
    PAWS-X & fr & 1{,}986 & 22{,}366 & $95.2_{\pm 9.8}$ & $81.3_{\pm 16.4}$  \\
    PAWS-X & es & 1{,}998 & 21{,}931 & $95.9_{\pm 8.5}$ & $84.9_{\pm 15.6}$  \\
    PAWS-X & zh & 1{,}975 & 20{,}204 & $92.4_{\pm 10.3}$ & $76.8_{\pm 16.4}$  \\
    PAWS-X & ja & 1{,}969 & 23{,}994 & $95.4_{\pm 8.8}$ & $82.0_{\pm 15.1}$ \\
    \bottomrule
    \end{tabular}
    \caption{Per-language question yield and content coverage,
    \texttt{qwen2.5-32b} teacher, uncapped banks. $N_{\text{ref}}$ counts
    references retaining at least one QC-admitted question; $|Q|$ is the
    resulting bank size. \emph{item} = question text + gold answer,
    \emph{answer} = gold answer alone; subscripts are standard deviations over
    references. }
    \label{tab:qayield}
        \end{table}
    
        \begin{table}[t]
        \centering\small
        \setlength{\tabcolsep}{3pt}

\begin{tabular}{lrrr}
\toprule
\textit{Content POS} & \textsc{num} & \textsc{adj} & \textsc{noun}  \\
\midrule
Phoenix-2014T (de) & 97.0 & 83.7 & 87.0 \\
CSL-Daily (zh)     & 93.1 & 96.6 & 90.7 \\
OpusParcus (de)    & 95.6 & 85.7 & 88.7 \\
PAWS-X (en)        & 92.3 & 83.8 & 83.3 \\
PAWS-X (zh)        & 88.6 & 84.5 & 82.6 \\
\midrule

\midrule
\textit{Function POS} & \textsc{pron} & \textsc{aux} & \textsc{sconj}\\
\midrule
Phoenix-2014T (de) & 35.7 & 10.5 & 33.3 \\
CSL-Daily (zh)     & 86.9 & --   & 46.4 \\
OpusParcus (de)    & 62.9 & 17.3 & 48.3 \\
PAWS-X (en)        & 45.8 & 16.1 & 19.9 \\
PAWS-X (zh)        & 61.1 & --   &  5.6 \\
\bottomrule
\end{tabular}
    \caption{Per-POS coverage by the gold answers (\%), representative banks.
    The metric's scoring mass sits on numerals, adjectives and nouns, and
    largely bypasses auxiliaries and subordinators. Pronouns are the exception
    and by design: a dedicated participant-extraction pass probes who acts and
    who is acted on, since a translation can reverse those and lose the sentence
    while keeping every noun in it, so pronoun coverage is high wherever the
    corpus is pronoun-dense (CSL-Daily $86.9$, OpusParcus $62.9$) and low only on
    the weather text of Phoenix-2014T, which contains few. Dashes are
    tags the language's tagger does not assign. }
    \label{tab:qaposcov}
        \end{table}

    
        \section{Per-language Validation Results}
        \label{app:validation}

        Table~\ref{tab:validation} of Section~\ref{sec:qavalidation} aggregates PAWS-X
        over languages; Table~\ref{tab:pawsx_flip} gives the breakdown. The QA advantage
        holds in all seven languages and is largest where BLEU-4 is weakest: $+0.171$ for
        English, where the \texttt{13a} tokeniser is well matched to the text, against
        $+0.250$ and $+0.255$ for Japanese and Korean, where character-level tokenisation
        leaves BLEU-4 near chance ($0.545$, $0.534$). The per-language OpusParcus
        correlations of Figure~\ref{fig:oprank} run from $\rho = 0.12$ for BLEU-4 against
        $0.53$ for QA in Russian to $0.40$ against $0.62$ in Swedish.
    
        \begin{table}[!h]
        \centering\small
        \setlength{\tabcolsep}{4pt}
        \begin{tabular}{lrrrr}
        \toprule
        \textbf{Lang.} & $n$ & \textbf{BLEU AUC} & \textbf{QA AUC} & $\Delta$ \\
        \midrule
        en & 1{,}995 & 0.705 & \textbf{0.875} & +0.171 \\
        de & 1{,}969 & 0.626 & \textbf{0.847} & +0.221 \\
        fr & 1{,}986 & 0.645 & \textbf{0.846} & +0.202 \\
        es & 1{,}998 & 0.629 & \textbf{0.850} & +0.221 \\
        zh & 1{,}975 & 0.594 & \textbf{0.818} & +0.224 \\
        ja & 1{,}969 & 0.545 & \textbf{0.795} & +0.250 \\
        ko & 1{,}970 & 0.534 & \textbf{0.790} & +0.255 \\
        \midrule
        mean & & 0.611 & \textbf{0.831} & +0.221 \\
        \bottomrule
        \end{tabular}
     
        \caption{PAWS-X meaning-flip discrimination (test split, \texttt{qwen2.5-32b} teacher). ROC-AUC = $P(\text{score(paraphrase)} > \text{score(flipped)})$ over the question-bearing pairs; $0.5$ = blind to the flip. Surface overlap is high in both classes by construction. QA wins in 7/7 languages.}
        \label{tab:pawsx_flip}
        \end{table}

      \section{Run-to-Run Stability: Sources and Per-System Numbers}
        \label{app:stability}

        Two mechanisms make a repeat of the pipeline differ, and only one of them is the
        serving stack. The second per-content-unit question-generation pass samples at
        $\tau=0.7$ by design, as a recall device for questions the greedy pass does not
        produce (Appendix~\ref{app:prompts}); no sampling seed is sent, so that pass
        essentially never repeats. Every other stage decodes greedily and is reproducible
        only up to the arithmetic of the server, where continuous batching varies batch
        composition and with it the floating-point reduction order. Comparing two runs
        over the calls they share, the greedy stages return identical text for $99.6$ and
        $99.7\%$ of content-unit extractions, $98.0$ and $98.3\%$ of distractor
        generations, and $99.96\%$ of answering calls on both benchmarks, the exposure
        scaling with output length. Question generation, which mixes the greedy and
        sampled passes, returns identical text for only $57.6$ and $62.8\%$ of its calls.
        Deliberate sampling is therefore the dominant channel and batching the secondary
        one, and their combined effect is that a repeat asks a substantially different
        bank: $71.7\%$ of admitted items recur between two runs on Phoenix-2014T and
        $77.9\%$ on CSL-Daily (Jaccard $0.56$ and $0.64$). Publishing the exact bank
        alongside a score is thus needed to reproduce a per-item analysis, but not to
        compare systems.

        \begin{figure}[!h]
            \centering
            \includegraphics[width=\linewidth]{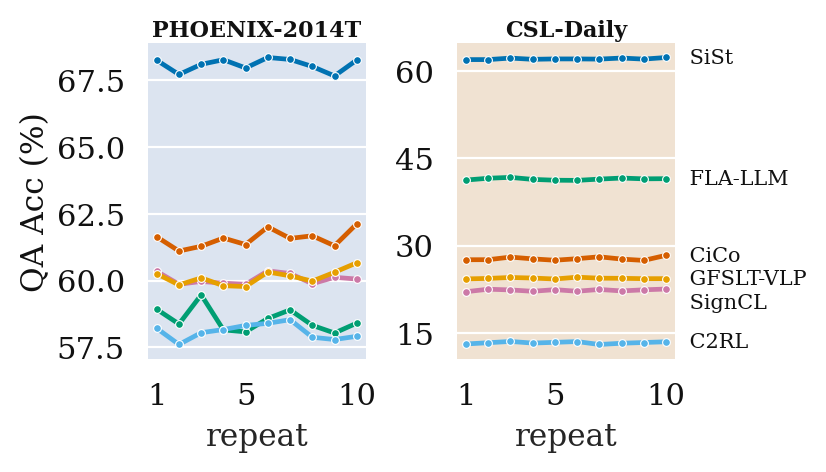}
            \caption{Per-system QA accuracy across the ten independent bank
            regenerations of Table~\ref{tab:stability}, one panel per benchmark in its
            identity colour. The y axes are independent (Phoenix-2014T spans thirteen
            points, CSL-Daily fifty). No system changes order in either panel.}
            \label{fig:qastability}
        \end{figure}

        \begin{figure*}[!t]  
            \centering
            \includegraphics[width=\linewidth]{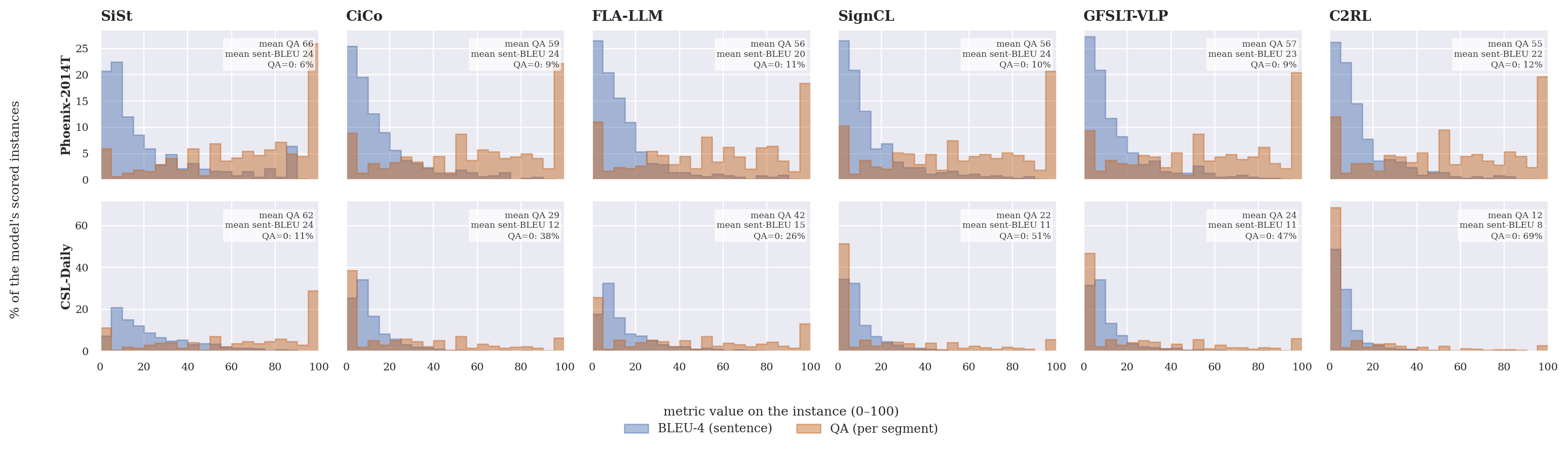}
            \caption{Per-instance distribution of sentence BLEU-4 and QA accuracy for each
            model, on Phoenix-2014T (top) and CSL-Daily (bottom).}
            \label{fig:performancedist}
        \end{figure*}

        The modal system ordering by QA is reproduced in every repeat on both benchmarks,
        a single distinct ordering with mean Kendall $\tau=1.00$ against it
        (Figure~\ref{fig:qastability}). This holds even for the closest pair,
        SignCL and GFSLT-VLP, whose means sit $0.35$ points apart on Phoenix-2014T, a gap
        of roughly one standard deviation of the difference between two repeats; their
        confidence intervals in Table~\ref{tab:combined_results} nevertheless overlap, so
        the pair is reproducibly ordered by the pipeline without being separated by the
        test set. BLEU-4 recomputed alongside is bit-identical across all ten
        Phoenix-2014T repeats, which admit the same $641$ references every time; on
        CSL-Daily the admitted set moves by a single reference in two of the ten repeats
        and BLEU-4 moves with it by at most $0.01$ points. The spread in QA therefore
        enters through which questions clear the gates, not through the scoring. 
        
        \section{Training-Likeness: Exclusions and Cumulative View}
        \label{app:overlap}
    
        \paragraph{Estimator.} Training-likeness uses the \texttt{fuzz.ratio}
        character-level similarity from RapidFuzz. Since $\ell_i$ is a property of the
        benchmark, all models are ordered identically. The sliding window spans $12$
        likeness points, stepped by $2$; inside each, the per-instance score is regressed
        on $\ell$ and on reference length, and the partial coefficient $\beta_m(\ell)$
        gives the sensitivity at fixed length, reported as
        \[
        S_m(\ell) = \frac{100 \, \beta_m(\ell)}{\bar C_m},
        \]
        with $\bar C_m$ model $m$'s mean per-instance score. The normaliser is sentence
        BLEU-4, not the corpus BLEU-4 reported in papers; the two differ by $1$--$2$
        points.

        \paragraph{Exclusions.} Two groups are excluded from the local sensitivity curve
        of Section~\ref{sec:sensitivity}. References at $\ell = 100$ are exact copies of
        a training sentence, a point mass that admits no slope; on Phoenix-2014T they are
        also half the length of their neighbours ($5.8$ against $12.5$ tokens), so a
        window straddling them would report a length difference as likeness sensitivity.
        These are reported as score levels instead. Windows holding fewer than $40$
        instances, or in which a single likeness value holds more than half the mass,
        cannot support a slope either and are left empty rather than interpolated.
    
        \begin{figure}[!h]
            \centering
            \includegraphics[width=\linewidth]{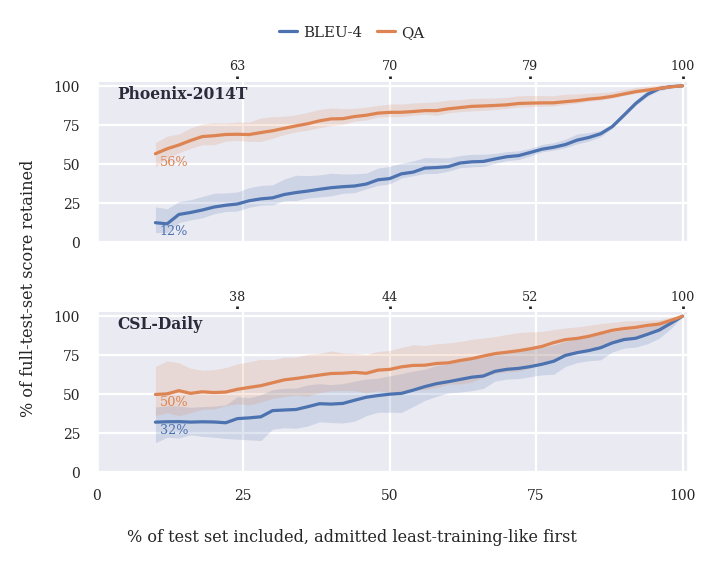}
            \caption{Retained fraction $R_m(f)=C_m(f)/C_m(1)$ over nested prefixes of the
            test set, admitted least-training-like first. Mean over $M=5$ models, band =
            min--max. Upper axis: training-likeness of the last admitted instance.
            Left-hand labels give $R_m$ at $f=0.1$.}
            \label{fig:retained}
        \end{figure}
    
              \paragraph{Cumulative view.} The local view of Section~\ref{sec:sensitivity} asks
        how fast a metric moves; this one asks how much of a published number the
        training-like instances account for in aggregate. We sort the test set from least
        to most training-like, score nested prefixes, and plot the retained fraction
        $R_m(f)=C_m(f)/C_m(1)$ (Figure~\ref{fig:retained}), normalising each model against
        itself before averaging so that differences in absolute level do not dominate. A
        metric insensitive to training overlap stays flat near $100\%$; one inflated by
        training-like instances rises steeply to the right. On Phoenix-2014T, BLEU-4
        retains $12.3\%$ of its reported score on the least training-like tenth of the
        test set against $56.5\%$ for QA; on CSL-Daily, $32.1\%$ and $49.8\%$.

        This view cannot replace the local one. $R_m$ is cumulative, and differentiating
        it does not recover $S_m$: for a running mean
        $\frac{d}{dk}\!\left[S(k)/k\right] = (x_k - C(k))/k$, so one local effect appears
        large early and small late purely through how many instances have already been
        averaged in, and corpus BLEU-4, not being a mean, adds a shifting n-gram
        normalisation on top. The two figures report the same phenomenon at different
        resolutions, and agree.

        \section{Per-instance Score Distributions}
        \label{app:perinstance}

        Section~\ref{sec:ranking} notes that the two metrics agree at corpus level while
        disagreeing on individual instances; Figure~\ref{fig:performancedist} shows the
        distributions behind that. Sentence BLEU-4 is massed near zero on both
        benchmarks, whereas QA occupies the whole range on Phoenix-2014T with its mode at
        $100$: on that benchmark a majority of instances are scored by BLEU-4 as
        near-total failures while QA reads much of their content as transferred. The
        share of instances at $\mathrm{QA}=0$, printed in each panel, is the clearest
        difference between the benchmarks: $7$--$10\%$ on Phoenix-2014T against
        $28$--$67\%$ on CSL-Daily, so the corpus means on CSL-Daily are driven
        substantially by instances from which nothing is recovered rather than by
        uniformly partial transfer.

\end{document}